\documentclass[10pt,twocolumn,letterpaper]{article}

\pdfoutput=1

\usepackage{times}
\usepackage{epsfig}
\usepackage{graphicx}
\usepackage{float}
\usepackage{wrapfig}
\usepackage{amsmath,amssymb,amsthm}
\usepackage{algorithm,algorithmicx,algpseudocode}
\usepackage{bm,xspace}
\usepackage{comment}
\usepackage{verbatim}
\usepackage{multirow}
\usepackage{balance}
\usepackage{url}
\usepackage{booktabs}
\usepackage{etoolbox,siunitx}
\usepackage{calc}
\usepackage{pifont,hologo}
\usepackage[usenames, dvipsnames]{color}
\usepackage{nicefrac}

\usepackage{epigraph}
\usepackage{enumitem}
\usepackage{soul}
\usepackage[small]{subfigure}
\usepackage{mdframed}
\usepackage{microtype}

\newcommand{\ra}[1]{\renewcommand{\arraystretch}{#1}}

\usepackage[super]{nth}
\usepackage{nicefrac}
\robustify\bfseries

\def\XX{\mathbf{X}}

\def\btheta{{\bm\theta}}

\DeclareMathOperator*{\argmax}{arg\,max}

\newcommand\ggreater{\mathbin{>\!\!\!>}}
\newcommand\llesser{\mathbin{<\!\!\!<}}

\DeclareMathSymbol{@}{\mathord}{letters}{"3B}

\def\latex/{\LaTeX}
\def\bibtex/{\hologo{BibTeX}}

\usepackage{iccv}
\usepackage{times}
\usepackage{epsfig}
\usepackage{graphicx}
\usepackage{amsmath}
\usepackage{amssymb}

\usepackage[accsupp]{axessibility}  

\usepackage{appendix}

\usepackage[breaklinks=true,bookmarks=false]{hyperref}

\iccvfinalcopy 

\def\iccvPaperID{3390} 

\ificcvfinal\fi

\begin{document}

\title{Online Continual Learning with Natural Distribution Shifts:\\ An Empirical Study with Visual Data}

\author{Zhipeng Cai\\
Intel Labs\\
\and
Ozan Sener\\
Intel Labs\\
\and
Vladlen Koltun\\
Intel Labs\\
}

\maketitle
\ificcvfinal\thispagestyle{empty}\fi

\begin{abstract}
Continual learning is the problem of learning and retaining knowledge through time over multiple tasks and environments. Research has primarily focused on the incremental classification setting, where new tasks/classes are added at discrete time intervals. Such an ``offline'' setting does not evaluate the ability of agents to learn effectively and efficiently, since an agent can perform multiple learning epochs without any time limitation when a task is added. We argue that ``online'' continual learning, where data is a single continuous stream without task boundaries, enables evaluating both information retention and online learning efficacy. In online continual learning, each incoming small batch of data is first used for testing and then added to the training set, making the problem truly online. Trained models are later evaluated on historical data to assess information retention. We introduce a new benchmark for online continual visual learning that exhibits large scale and natural distribution shifts. Through a large-scale analysis, we identify critical and previously unobserved phenomena of gradient-based optimization in continual learning, and propose effective strategies for improving gradient-based online continual learning with real data. The source code and dataset are available in: \url{https://github.com/IntelLabs/continuallearning}.
\end{abstract}

\section{Introduction}\label{sec:intro}

Supervised learning aims to find models that can predict labels given
input data, with satisfactory performance when evaluated on a specific population of interest.
This population is typically sampled to create training data for the model to learn over. The critical requirement for successful learning is a set of training data
points that are independent and identically distributed (iid.). Continual learning challenges this
assumption and considers a sequence of learning problems where the distribution changes dramatically through
time. This setting is crucial when learned models are deployed in interactive systems, since the environment with which agents interact continually evolves.

There are two critical performance metrics for a continual learner: learning efficacy and information
retention. Learning efficacy entails the simple question, ``is learning the n-th thing easier than
learning the first thing?''~\cite{thrun1995}. This ability is critical for fast
learning and quick adaptation. Information retention considers the model's ability
to quickly recall when faced with a historical task that had been previously considered. This question is also
studied to understand a property of neural networks called catastrophic forgetting~\cite{mccloskey1989catastrophic}.

Continual learning algorithms are typically evaluated in an incremental classification setting, where tasks/classes arrive
one-by-one at discrete time intervals. Multiple learning epochs over the current task are permitted and the learner can spend as much time as desired to learn each task. This setting
is appropriate for evaluating information retention, because access to
previous tasks is prohibited. However, learning efficacy is not evaluated in this setting, because models can easily
learn each task from scratch and still be successful~\cite{prabhu2020gdumb}. We refer to this incremental classification setting as \emph{offline} continual learning.

Our work focuses on \emph{online} continual learning, which aims to evaluate learning efficacy in addition to information retention.
In online continual learning, there is a single online stream of data.
At every time step, a small batch of data arrives. The model needs to immediately predict labels
for the incoming data as an \emph{online testing step}. After the prediction, this batch of data is
added to the dataset. The model needs to be updated before the next batch of data arrives using a
fixed budget of computation and memory; this is an \emph{online training step}. Testing and
training are on the fly. Moreover, as we are in a continual learning setting, the data distribution changes over time. A successful online test performance requires efficient learning and adaptation in this non-stationary setting.

To study online continual visual learning, we construct a new benchmark where the data distribution
evolves naturally over time. To do so, we leverage images with geolocation tags and time stamps.
We use a subset of YFCC100M~\cite{thomee2016yfcc100m} with 39 million images captured over 9 years.
Our task is \emph{online continual geolocalization}.
We empirically evaluate the natural distribution shift and
validate that the benchmark is appropriate to study online continual learning. We further use this
benchmark to analyze the behavior of gradient-based optimization in online continual learning.
Our experiments suggest that the non-stationarity in the data results in a significantly different behavior from what was previously observed in offline continual learning. Surprisingly, learning efficacy and information retention
turn out to be conflicting objectives from an optimization perspective, necessitating a careful trade-off. We also found that mini-batching is a non-trivial problem for online continual learning. Increasing batch sizes in SGD, even by a small factor, significantly hurts both learning efficacy and information retention. Based on the analysis, we propose simple yet effective strategies, such as online learning rate and replay buffer size adaptation algorithms, that significantly improve gradient-based optimization for online continual learning. We will share our benchmark and code with the community in order to support future research in online continual visual learning.

\section{Related Work}\label{sec:Related}

\paragraph{Continual learning benchmarks.}
Most visual continual learning benchmarks use synthetic task sequences. A common way to
synthesize task sequences is to separate the original label space into multiple subspaces. Split
MNIST~\cite{zenke2017continual}, Split notMNIST~\cite{nguyen2018variational}, Split
CIFAR10/100~\cite{zenke2017continual}, Split Tiny-ImageNet~\cite{chaudhry2019tiny}, and
iILSVRC~\cite{rebuffi2017icarl} are all constructed in this way. Another approach is to inject different types of transformations into data as new tasks.
Permuted MNIST~\cite{zenke2017continual} and Rotated MNIST~\cite{lopez2017gradient} are constructed in this way.
For online visual continual learning, Aljundi et al.~\cite{aljundi2019task} use soap opera series for actor identification. Though training is online, testing is done once per episode, on time-invariant validation data from all episodes. 
Recently, the Firehose benchmark~\cite{hu2020drinking} proposed online continual learning of natural language models.
It contains a large stream of Twitter posts used to train a self-supervised and multi-task language model for per-user tweet prediction.

\vspace{-1em}
\paragraph{Continual learning algorithms.}
Existing continual learning algorithms can be roughly categorized into 1) regularization-based, 2) parameter-isolation-based, and 3) replay-based (see~\cite{de2019continual} for a detailed survey). Regularization-based methods add regularization terms into the training loss, based on distillation~\cite{li2017learning} or estimated network parameter importance~\cite{kirkpatrick2017overcoming, zenke2017continual, aljundi2018memory, lee2017overcoming}. Since historical data are not cached for training, these methods often suffer from catastrophic forgetting, especially given a long data stream~\cite{farquhar2018towards}. Parameter-isolation-based methods~\cite{mallya2018packnet, serra2018overcoming} assign different sets of network parameters to different tasks. However, they are not suitable for online continual learning, due to the use of task ID for training. Replay-based methods train models using historical examples~\cite{rebuffi2017icarl, chaudhry2019tiny} or examples synthesized by generative models that are trained on historical samples~\cite{shin2017continual}. The replay data can serve as part of the training data, or be utilized to constrain the gradient direction during training~\cite{lopez2017gradient, chaudhry2018efficient}. Replay-based methods have generally been found more effective than the other methods and we adopt them as our algorithmic starting point in this work.

\vspace{-1em}
\paragraph{Geolocalization algorithms.}
We use geolocalization on a stream of images as an online continual learning setting. Though our work is the
only continual/online approach, the problem of geolocalization has been widely studied. The pioneering work of
Hays and Efros~\cite{hays2008im2gps, hays2015large} addresses geolocalization by image retrieval. Given a query image,
 it performs a nearest neighbor search on millions of images with geolocation tags and uses the
 location of the nearest neighbour for prediction. Vo et al.~\cite{vo2017revisiting} replace handcrafted retrieval features with ones computed by a deep neural network.
 This network is trained with a classification loss but only used for feature extraction.
 PlaNet~\cite{weyand2016planet} formulates geolocalization into an image classification problem
 and trains a deep neural network for classification. PlaNet is much more efficient than retrieval-based approaches because it only
 requires one forward pass per query image, rather than nearest neighbor searching over millions of images.
 CPlaNet~\cite{seo2018cplanet} introduced combinatorial partitioning to PlaNet, which generates fine-grained class labels
 using the intersection of multiple coarse-grained partitions of the map. This technique is used to alleviate the conflict
 between the large label space and the small number of training examples per class. We use PlaNet as a starting point due to its
 simplicity and efficiency, and extend it to continual and online settings.

\section{Online Continual Learning}\label{sec:OCL}
In this section, we formally define online continual learning (OCL), mainly following the
definition of online learning. We further discuss the metrics we use to evaluate learning
efficacy and information retention.

Following the common definition of online learning \cite{hazan2019introduction, shalev2011online}, we
define OCL as a game between a learner and an environment. The
learner operates in a predefined function class $h(\cdot;\btheta)\colon \mathcal{X} \rightarrow \mathcal{Y}$
 with parameters $\btheta$ predicting the label $Y \in \mathcal{Y}$ given an input $\XX \in \mathcal{X}$. At each step ${t \in \{1,2,...,\infty\}}$ of the game,
\begin{itemize}[noitemsep,topsep=2pt,parsep=2pt,partopsep=2pt]
    \item The environment generates a set of data points \mbox{$\XX_t \sim \pi_t$} sampled from a
    non-stationary distribution $\pi_t$.
    \item The learner makes predictions for $\XX_t$ using the current model $\btheta_t$ as
    $\hat{Y}_t = h(\XX_t; \btheta_t)$.
    \item The environment reveals the true labels $Y_t$ and computes the online metric
    $m(Y_t, \hat{Y}_t)$.
    \item The learner updates the model $\btheta_{t+1}$
    using a fixed budget of computation and memory.
\end{itemize}

Since this formalization directly follows online learning, it is only designed for evaluating
online performance. To further assess information retention,
we evaluate the models $\btheta_{\nicefrac{H}{3}}, \btheta_{\nicefrac{2H}{3}}, \btheta_{H}$
at predefined time instants $\nicefrac{H}{3}$, $\nicefrac{2H}{3}$, $H$, where $H$ is the total number of time steps for the evaluation. The models are evaluated over validation data (not seen during training) sampled at historical time steps. We refer to this as \emph{backward transfer}. Similarly, we also evaluate the models on future data and call this \emph{forward transfer}.

\vspace{-1em}

\paragraph{Metrics.}
To evaluate learning efficacy, we measure the \emph{average online accuracy} over time,
similar to average regret in online learning. We compute the running average of the computed accuracy on the fly as
\begin{align}\label{eq:online_fit}
    & \texttt{acc}_{\texttt{O}}(t) = \frac{1}{t} \sum_{s=1}^t{\texttt{acc}(Y_s, \hat{Y}_s)}.
\end{align}

In order to evaluate information retention, we compute \emph{backward transfer} at various predefined time
instants. We specifically compute the average accuracy in the historical data. Formally, the
backward transfer for the model at time $T$ ($T$ is $\nicefrac{H}{3}$, $\nicefrac{2H}{3}$,
or $H$ for us) is defined as
\begin{align}
    \texttt{acc}_{\texttt{B}}\texttt{@T}(t_B) = \frac{1}{t_B} \sum_{s=T-t_B}^T \texttt{acc}(Y_s,h(\XX_s; \btheta_T)).
\end{align}

Finally, we also evaluate how well the model transfers information to the future. This metric can be used to evaluate out-of-distribution generalization of the model through time.
Similar to backward transfer, we evaluate \emph{forward transfer} at a specific time point $T$
(specifically $\nicefrac{H}{3}$ and $\nicefrac{2H}{3}$) as
\begin{align}
    \texttt{acc}_{\texttt{F}}\texttt{@T}(t_F) = \frac{1}{t_F} \sum_{s=T}^{T+t_F} \texttt{acc}(Y_s,h(\XX_s; \btheta_T)).
\end{align}

All three metrics are a function of time, not a single time-invariant value. This choice is intentional
as the problem of interest is online. When we compare two algorithms, if one works better at earlier times and the other works better in later times, we would like to detect these characteristics.
We thus plot these three metrics through time for evaluation.

\section{Continual Localization Benchmark}\label{sec:CLOC}

\begin{figure}[t]
	\center
	\subfigure[S2 Cells in our dataset]{\includegraphics[width=0.9\columnwidth, height=0.5\columnwidth]{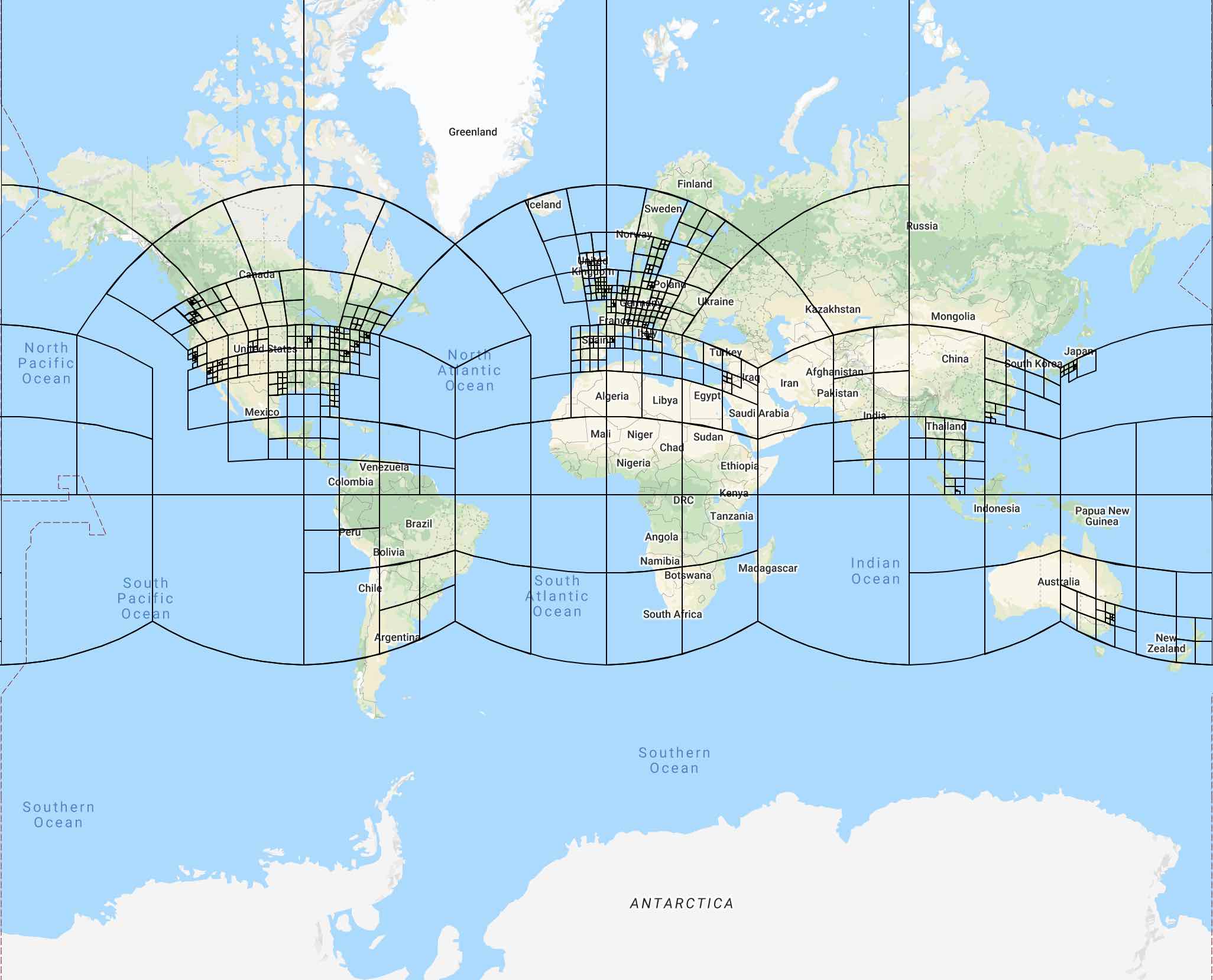}\label{fig:S2cell}}
	\subfigure[Distribution of number of images per country]{\includegraphics[width=0.9\columnwidth]{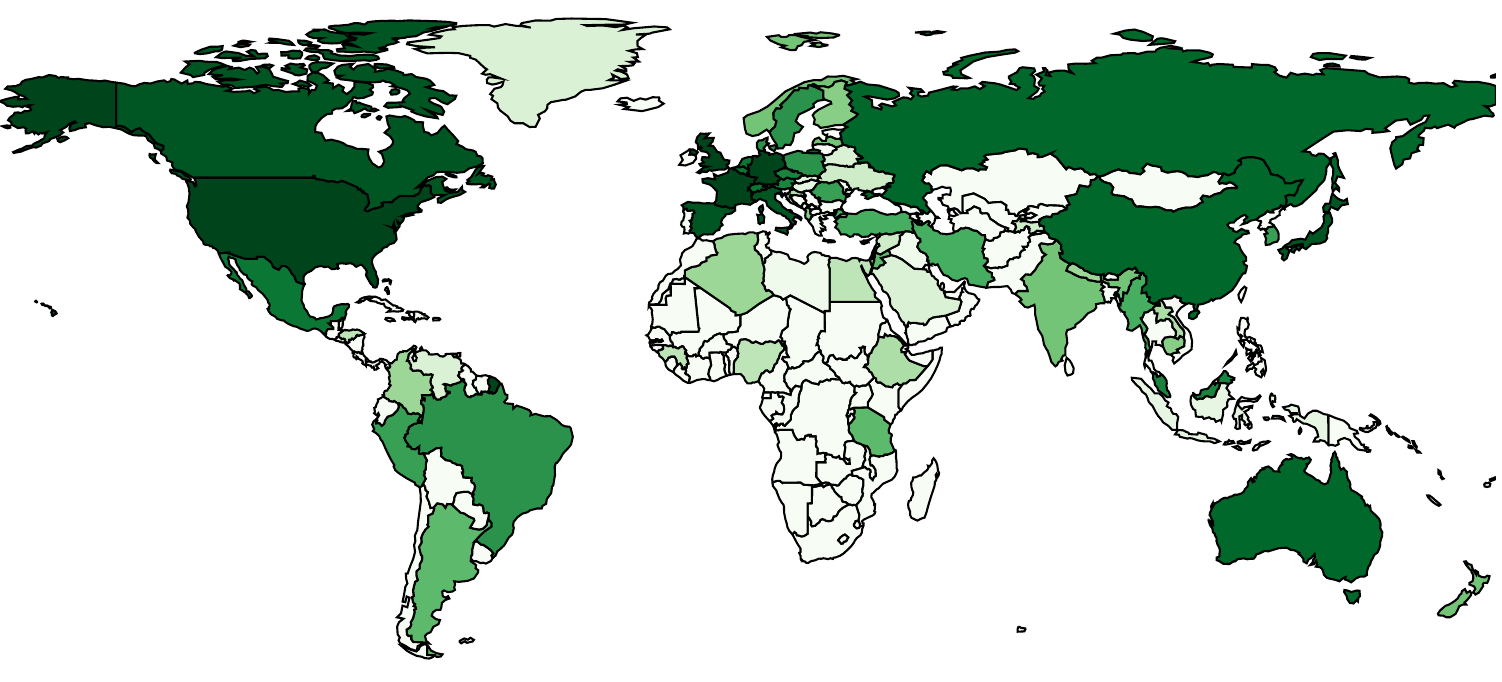}\label{fig:DisCountry}}
	\subfigure[Statistics over years]{\includegraphics[width=0.485\columnwidth]{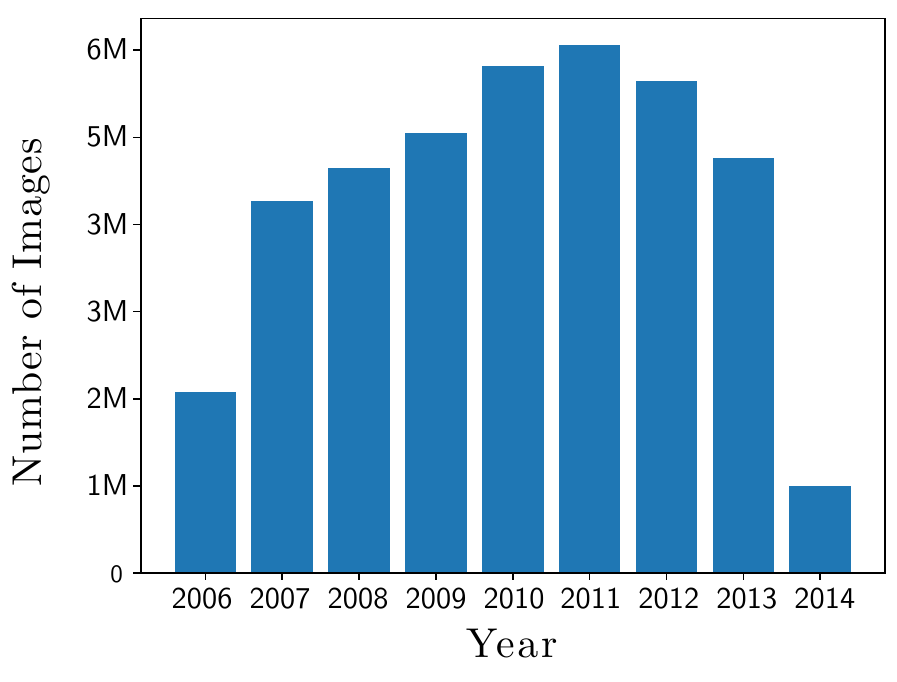}\label{fig:NOvsYear}}
\subfigure[Geolocation accuracy]{\includegraphics[width=0.485\columnwidth]{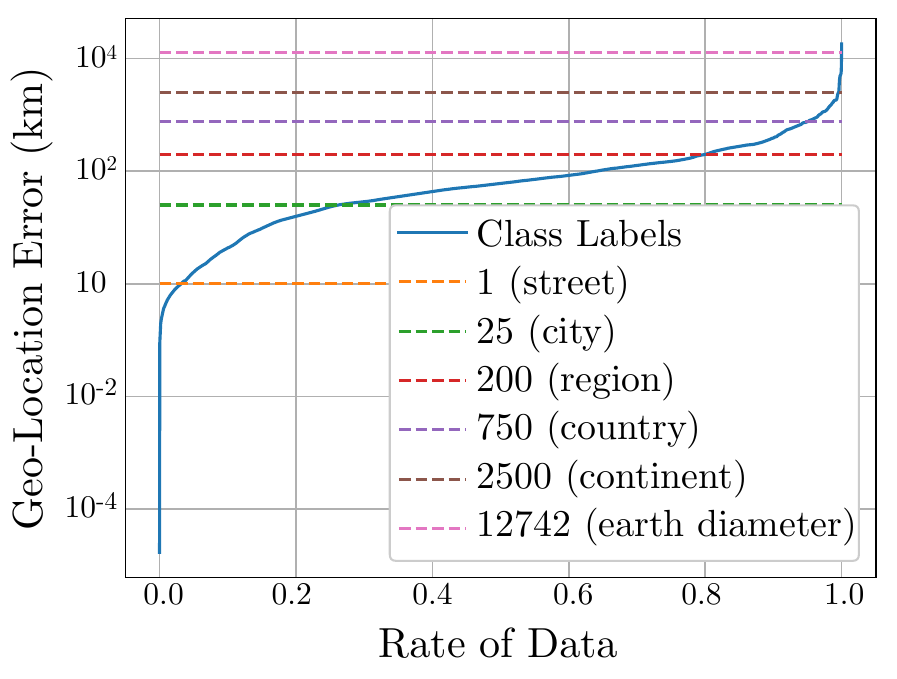}\label{fig:LabelAcc}}

	\caption{\textbf{Statistics of our benchmark}. (a) We convert the geolocalization problem into
		classification over 712 S2 Cells, visualized here. (b) We plot the distribution of
		images over countries. Dominating number of images come North America and Europe. (c) Number of images
		per year changes over time. (d) Geolocalization accuracy of the S2 Cells. (Lower right is better.)
		Following~\cite{hays2015large}, approximate scales of streets, cities, regions,
		countries and continents are plotted.}\label{fig:CLOC_stats}
	\vspace{-1.5em}

\end{figure}

To study online continual visual learning, we need a large-scale benchmark with natural distribution shifts.
We propose using geolocalization as a task for online continual visual learning, since the data is readily
available. We use images with their geolocations and time stamps from YFCC100M~\cite{thomee2016yfcc100m}.

After obtaining time stamps for YFCC100M images, we perform a series of pre-processing steps. First, we
keep only images from 2004 to 2014 since the images from pre-2004 typically have noisy time stamps in the
 form of the default date and time of the users' physical camera. Then, we order the data stream according to the time stamp, i.e., images taken earlier will be seen by the continual learner first. Finally, we allocate the first 5\% of the data
 for offline preprocessing and randomly sample another 1\% over the entire time range as the held-out validation
 set to evaluate backward and forward transfer. We have 39 million images for continual learning after this preprocessing,
 2 million images for offline preprocessing, and 392 thousand images for backward and forward transfer evaluation. We name our
 benchmark CLOC, short for Continual LOCalization. We visualize the distribution of these images over years
 and countries in Figure~\ref{fig:CLOC_stats}. We can see from the figure that the distribution of the images is
 heavily biased towards North America and Europe due to the distribution of Flickr users.

 We apply a standard procedure to convert the geolocalization problem into classification~\cite{weyand2016planet}. Specifically, we
 divide the earth into multiple regions using the
 S2 Cell hierarchy~\cite{url:s2cell}, following statistics computed on the offline preprocessing set.
 We also limit the number of images within each class to be between 50 and 10K. Using this process,
 we generate 712 S2 cells over the earth, converting our problem into a 712-class classification problem. We visualize
 the S2 cells in Figure~\ref{fig:S2cell}. Regions with more images are subdivided more finely.

In order to quantify the loss of geolocalization accuracy due to the constructed class labels, we plot in Fig.~\ref{fig:LabelAcc} the distribution of distances between the location of each image and the corresponding class center, i.e., the center of the S2 Cell. We can see that roughly
  $5\%$ of the data are $<1km$ from their class center at the street level. Roughly $30\%$ of the data
  are $<25km$ from their class center at the city level.  Roughly $80\%$ of the data are $<200km$ from their
  class center at the region level.

Compared to previous visual continual learning benchmarks, CLOC has a much larger scale: it contains 39 million images taken over 9 years, with 712 classes. Crucially, the distribution shift in CLOC is natural rather than synthetic. CLOC requires neither explicit task boundaries nor artificially injected data augmentations. The data distribution naturally evolves over time.

To perform OCL on CLOC, we ingest at each time step the consecutive set of images uploaded by the same user, since users sometimes upload at the same time a small album of images, which are typically from the same region. Once
the images arrive, the model needs to make predictions, update its model, and move to the next set of images. Note that images are ordered with the time stamps. Hence the OCL model receives these images in their natural order. 

\begin{figure}\center
	\subfigure{\includegraphics[width=1.0\columnwidth]{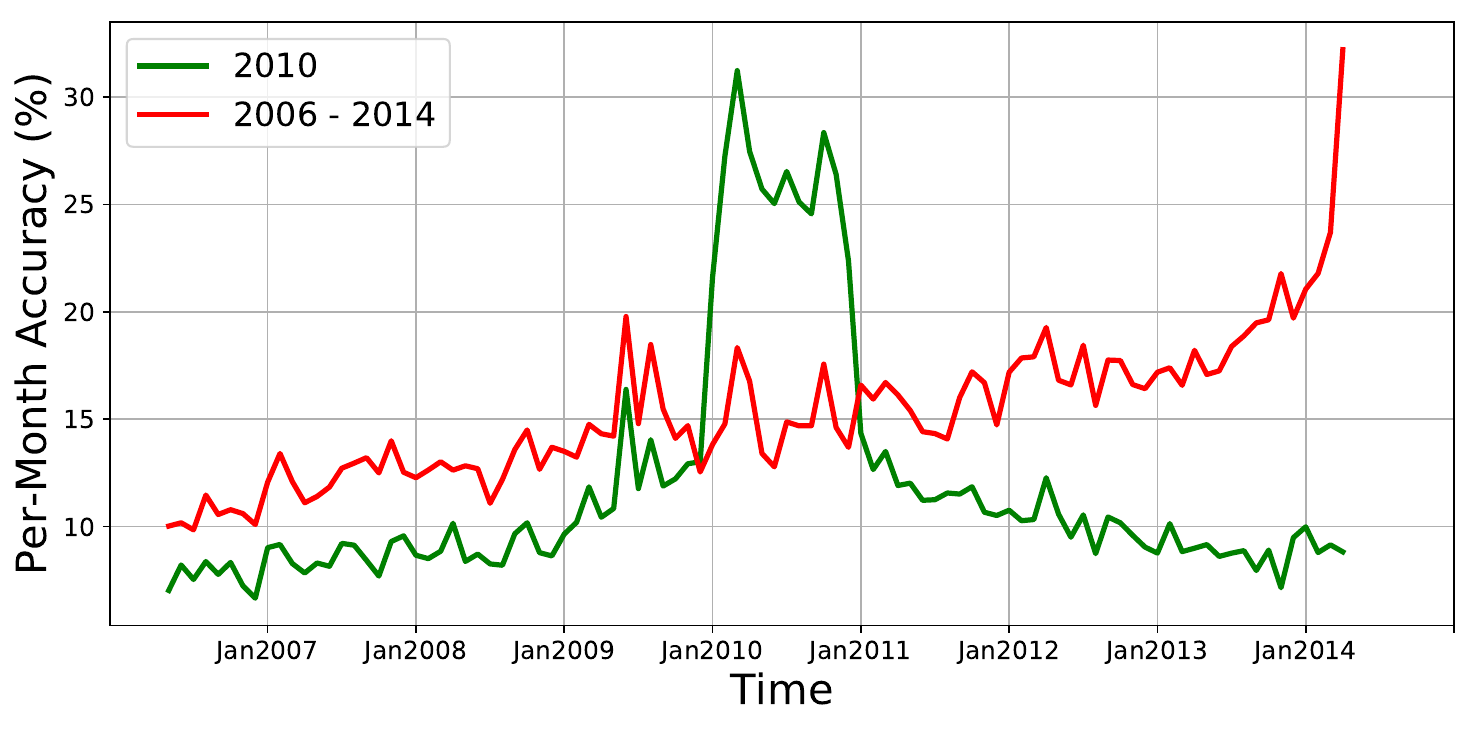}}
	\caption{\textbf{Distribution shift in CLOC.}
		We train two supervised models, one using data from the entire temporal range and the other only on data from the year 2010.
		We evaluate both models on the full temporal range using the validation set (not seen during training). Due to non-stationarity in the data, the performance of the 2010 model drops sharply on data from other times.}\label{fig:CLOC_disShift}
	\vspace{-1.5em}
\end{figure}

\vspace{-1em}

\paragraph{Validating the Distribution Shift of CLOC.} One key question that we need to validate is the
continual nature of the problem. If the data distribution is not changing through time, there is no need for
continual learning. To validate the distribution shift, we train two supervised learning models on CLOC (see Section~\ref{sec:Baseline} for
implementation details). The first model is trained over the entire temporal range, and the second only on data from the year 2010. For fair comparison, we subsample the training set of the full-range model to the same size as the training set of the other model. We evaluate the trained models on the held-out validation data and plot the top-1 accuracy vs.\ time in Figure~\ref{fig:CLOC_disShift}. The performance of the temporally localized model drops significantly on data from other times. In contrast, the full-range model does not have a sharp performance increase in year 2010. This indicates that there is significant non-stationarity in CLOC.

\section{Empirical Study and Results}
\subsection{Setup}
\label{sec:Baseline}

To empirically study online continual visual learning, we make a series of systematic choices focusing on simplicity,
scalability, and reproducibility. Our model predicts the label for each image independently, though sometimes multiple images from the same user may be tested together. We choose the ResNet50~\cite{he2016deep} architecture and the experience replay
(ER)~\cite{chaudhry2019tiny} algorithm as the main setting. We choose ER since it is an effective continual learning algorithm that scales to large-scale problems.
ER uses a \emph{replay buffer} to store a set of historical data. The strategy for keeping the replay buffer is important for continual learning. However,
our analysis suggests that it has a minor impact on gradient-based optimization, which is the main focus of our work. Hence,
we deploy the First-In-First-Out (FIFO) buffer for analysis in the main paper and empirically evaluate this choice in the supplement.

In each training iteration, images from the data stream are joined together with images randomly sampled from the replay
buffer to form the training batch. Note that the learning efficacy, i.e., eq.~\eqref{eq:online_fit}, is measured only on the data stream, without including the replay data.
We apply standard data augmentations to images~\cite{he2016deep}.
Augmentations are applied to both streaming and replay data. In other words, the replay buffer contains original images, and a random augmentation is applied whenever an image is replayed.

The concept of batch size in OCL does not exist as models need to predict the label for each incoming image online. In its direct implementation, this corresponds to a ``batch size'' of $1$ album, i.e., the next consecutive set of images uploaded by the same user (1.157 images on average).
However, analysis from supervised learning suggests that mini-batching is beneficial for gradient-based optimization~\cite{batch_size_analysis};
moreover, computational considerations also suggest some form of mini-batching since performing $\frac{39\text{ million}}{1.157} \approx 34\text{ million}$ iterations in order to test a model would be prohibitively costly.
Thus, we relax this constraint while keeping the online nature of the problem. The model receives $256$ images at each
iteration, but the online accuracy is computed over only the first album in these images.

Choosing hyperparameters for continual learning is especially tricky since the continual setting precludes ``rerunning'' models in identical conditions for hyperparameter selection.
Once a model is deployed, it needs to learn continually without retraining. Hence,
we make all hyperparameter decisions using the offline preprocessing dataset. The hyperparameters set by this method are the initial learning rate ($0.05$), weight decay ($1\mathrm{e}{-4}$), and loss (cross-entropy). We also report offline supervised learning results as a baseline, with hyperparameters  tuned separately on the whole dataset. As a result, the supervised learning baseline uses an initial learning rate of $0.025$, weight decay of  $1\mathrm{e}{-4}$, the cross-entropy loss, and the cosine learning rate schedule.

	\begin{figure}[t]
	\center
	\subfigure[Learning rate]{\includegraphics[width=0.95\columnwidth]{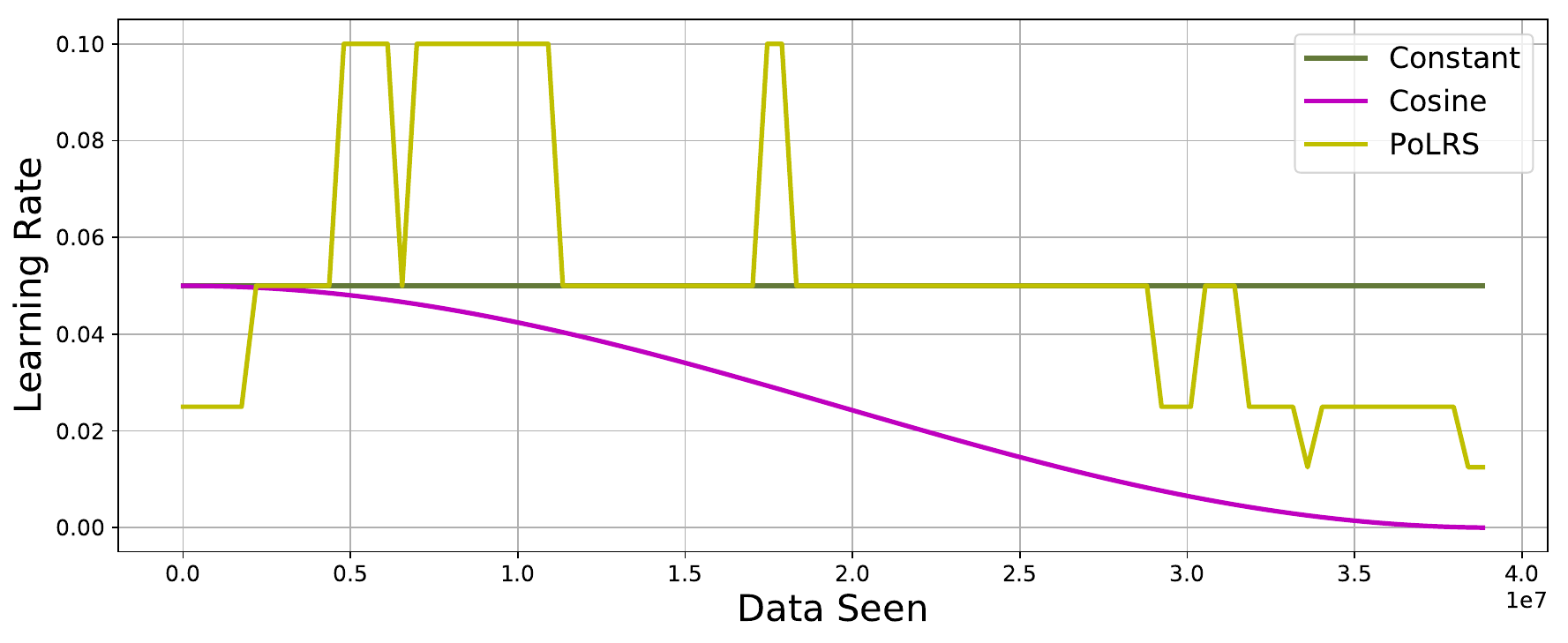}\label{fig:LR_over_time}}
	\subfigure[Average online accuracy ($\uparrow$)]{\includegraphics[width=0.49\columnwidth]{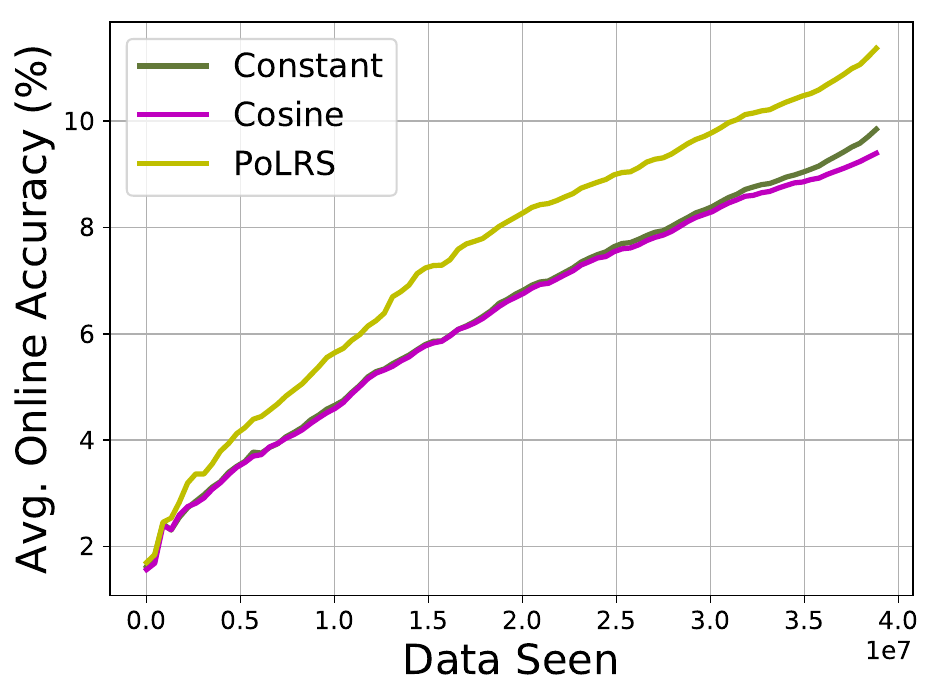}\label{fig:online_fit}}
	\subfigure[Backward transfer at $H$ ($\uparrow$)]{\includegraphics[width=0.49\columnwidth]{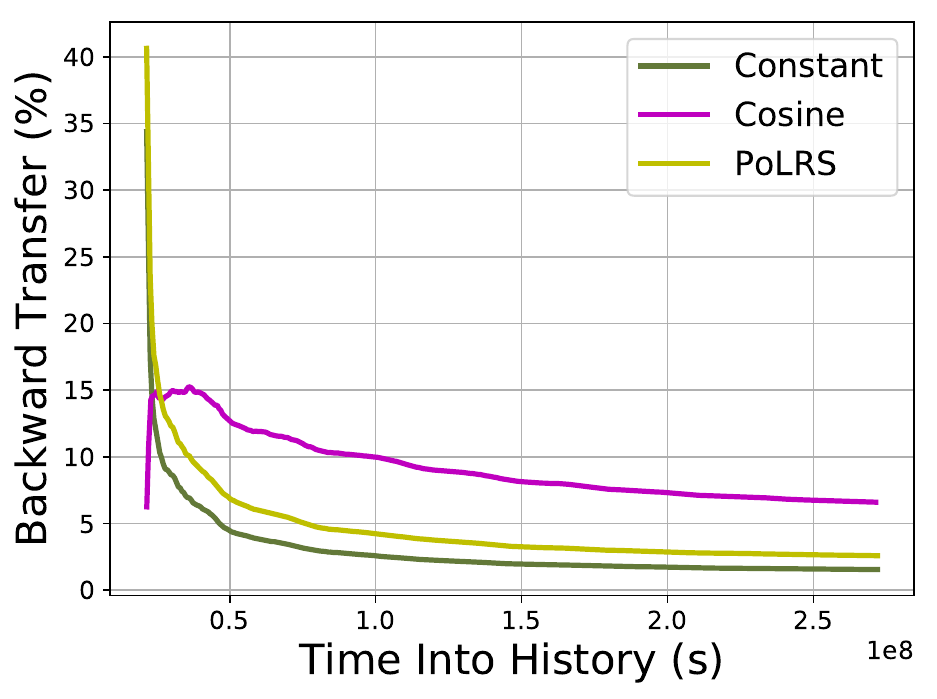}\label{fig:BT_epoch90}}
		\subfigure[Backward transfer at $\nicefrac{2H}{3}$ ($\uparrow$) ]{\includegraphics[width=0.49\columnwidth]{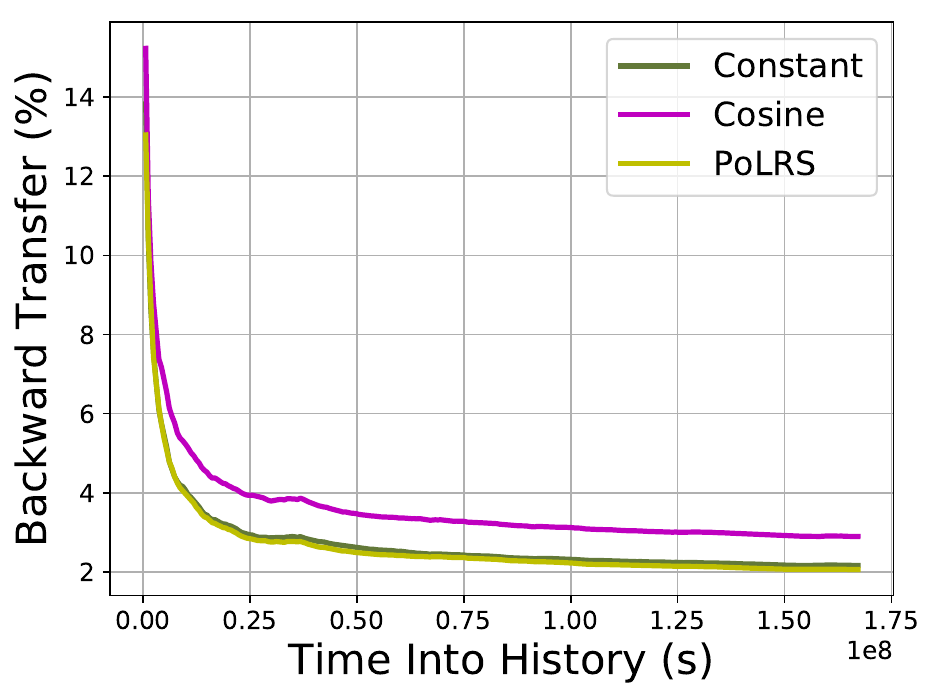}\label{fig:BT_epoch60}}
	\subfigure[Forward transfer at $\frac{2H}{3}$ ($\uparrow$)]{\includegraphics[width=0.49\columnwidth]{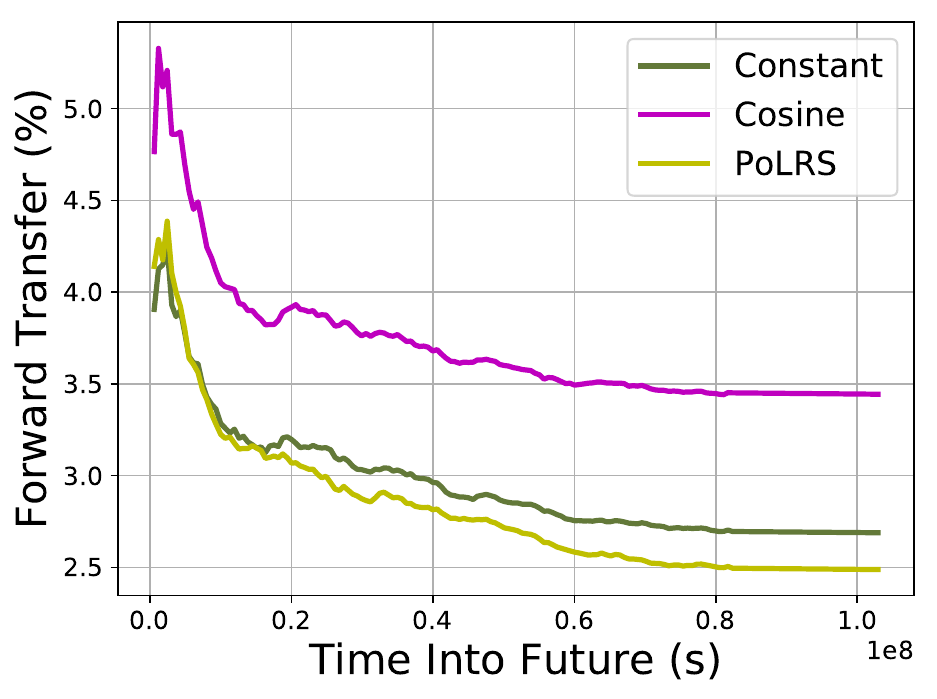}\label{fig:FT_epoch60}}
	\subfigure[Backward transfer at $\nicefrac{H}{3}$ ($\uparrow$)]{\includegraphics[width=0.49\columnwidth]{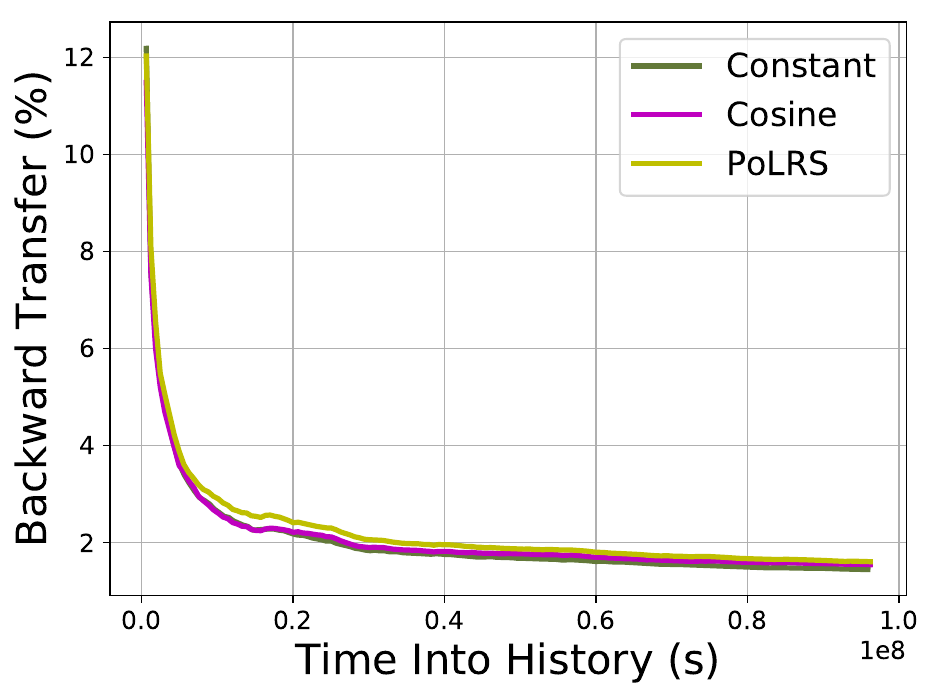}\label{fig:BT_epoch30}}
	\subfigure[Forward transfer at $\nicefrac{H}{3}$ ($\uparrow$)]{\includegraphics[width=0.49\columnwidth]{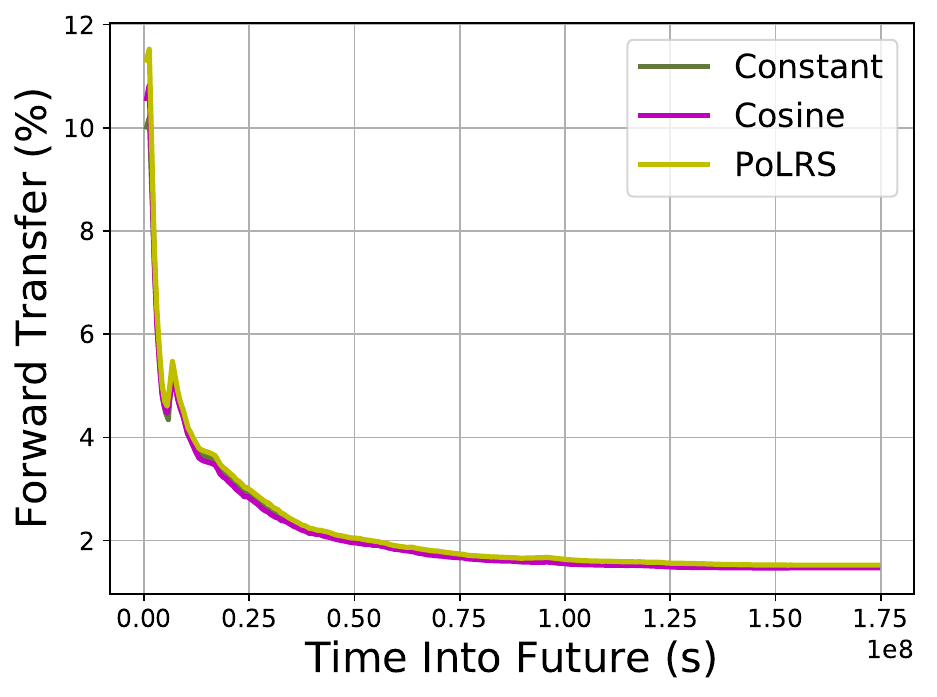}\label{fig:FT_epoch30}}
	\caption{\textbf{Learning rate analysis results}. $H$ is the total number of time steps. Arrows in the subfigure captions point towards better performance. (a) PoLRS adjusted the learning rate dynamically over time. (b) PoLRS out-performed fixed schedules in terms of average online accuracy. (c) The cosine schedule had the best backward transfer at time $H$, though its average online accuracy was the worst. (d,e) The cosine schedule had the best transfer when its learning rates were much smaller then other schedules ($\nicefrac{2H}{3}$ and $H$). (f,g) Schedules using large learning rates tend to have low forward and backward transfer.}
	\vspace{-1.5em}
\end{figure}

\subsection{Learning Rate Analysis}

In order to study the role of learning rates and their schedule in OCL, we compare a constant learning rate schedule where
the learning rate is fixed through time, and the cosine schedule~\cite{loshchilov2016sgdr} with one cycle and minimum LR of
0. We also evaluate an adaptive schedule generated by population-based search~\cite{jaderberg2017population}. Population-based search trains 3 models in parallel (keeping the total amount of computing the same as in the other conditions); every $N^{th}$ step, the weights of the best-performing model are copied to the others, and the learning rates of all models are centered at the current best learning rate. We set $N$ to 2 million, which is roughly the same size as the offline preprocessing step used to set initial hyperparameters. We summarize this schedule in Alg.~\ref{alg:OGS}, and call it PoLRS, short for Population Learning Rate Search.

	\begin{algorithm}
		\begin{algorithmic}[1]
			\Require Learning rate update interval $N$, metric for performance evaluation $m(\cdot)$,
			initial learning rate $l_0$, initial models $\{\btheta_1^1, \btheta_1^2, \btheta_1^3\}$.
			\State $j^* \leftarrow 1$ and set learning rates to $\{2l_0, l_0, 0.5l_0\}$.
			\For{$t \in \{1,2,...\}$}
			\State Test on $\XX_t$ using $\btheta_t^{j^*}$ and compute cost $m_t()$
			\State Update $\btheta_t^1, \btheta_t^2, \btheta_t^3$ and their metrics $m_t^1, m_t^2, m_t^3$
			\State  $j^* = \argmax_j(m_t^j)$\label{line:j*}
			\If{$t\mod N = 0$}
			\State Copy $\btheta_t^{j^*}$ to $\btheta_t^1, \btheta_t^2, \btheta_t^3$.
			\State Set learning rates to $\{2l_t^{j^*}, l_t^{j^*}, 0.5l_t^{j^*}\}$
			\EndIf
			\EndFor
		\end{algorithmic}
		\caption{Population Learning Rate Search (PoLRS)}\label{alg:OGS}
	\end{algorithm}

    We plot the learning rate of each method over time in Fig.~\ref{fig:LR_over_time}.
    PoLRS had a significantly different learning rate from other methods. We used the online accuracy as the
	metric while using PoLRS, and the online accuracy results from Fig.~\ref{fig:online_fit} show that PoLRS outperformed
    the other schedules. We further evaluated
	forward and backward transfer. Fig.~\ref{fig:BT_epoch90} shows the backward transfer of all schedules at the final 
	time step $H$. The cosine schedule had the best backward
	transfer but the worst online fit. Hence, the best learning rates for learning efficacy and information retention are different.
    Figs.~\ref{fig:BT_epoch60} to~\ref{fig:FT_epoch30} further plot the forward and backward transfer
     at times $\nicefrac{H}{3}$ and $\nicefrac{2H}{3}$.  We see that all schedules had low forward and backward transfer when their
	 learning rates were large, i.e., at time $\nicefrac{H}{3}$. Moreover, when the learning rate of the cosine schedule became
	 much smaller than the others, i.e., at time $\nicefrac{2H}{3}$ and $H$, it consistently outperformed the other schedules.
	 These results clearly show that small learning rates facilitate transfer, though potentially at the cost of learning efficacy.

\vspace{-1.5em}

\paragraph{Summary of findings.} (1) The ideal learning rates for learning efficacy and transfer are radically
	 different. (2) If \emph{learning efficacy} is prioritized, adaptive learning rate schedules, e.g., produced by PoLRS,
	 are better. (3) If \emph{transfer} is prioritized, and there is a pre-defined learning horizon $H$,
	 the cosine schedule is a good choice.

\subsection{Replay Buffer Size Analysis}

In order to analyze the impact of replay buffer sizes, we train three models with replay buffer sizes of 40 thousand, 4 million, and 39 million,
respectively. We use the cosine learning rate schedule for this analysis.


\begin{figure}
	\center
	\subfigure[Average online accuracy ($\uparrow$)]{\includegraphics[width=0.55\columnwidth]{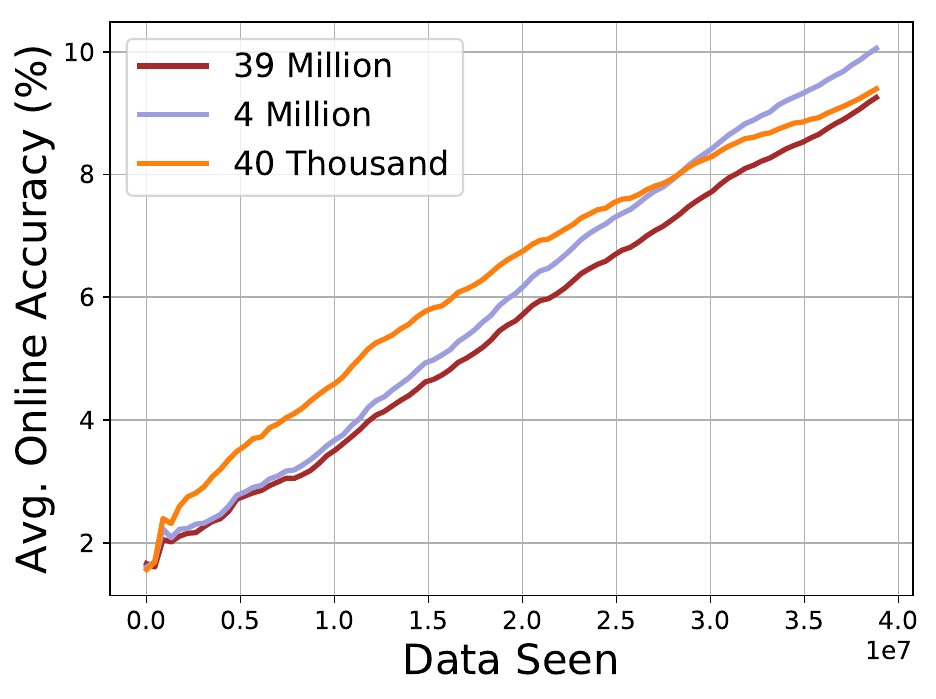}\label{fig:varied_buf_size}}
	\subfigure[Training accuracy $\texttt{acc}_{\texttt{Stream}}$ on the data stream]{\includegraphics[width=0.49\columnwidth]{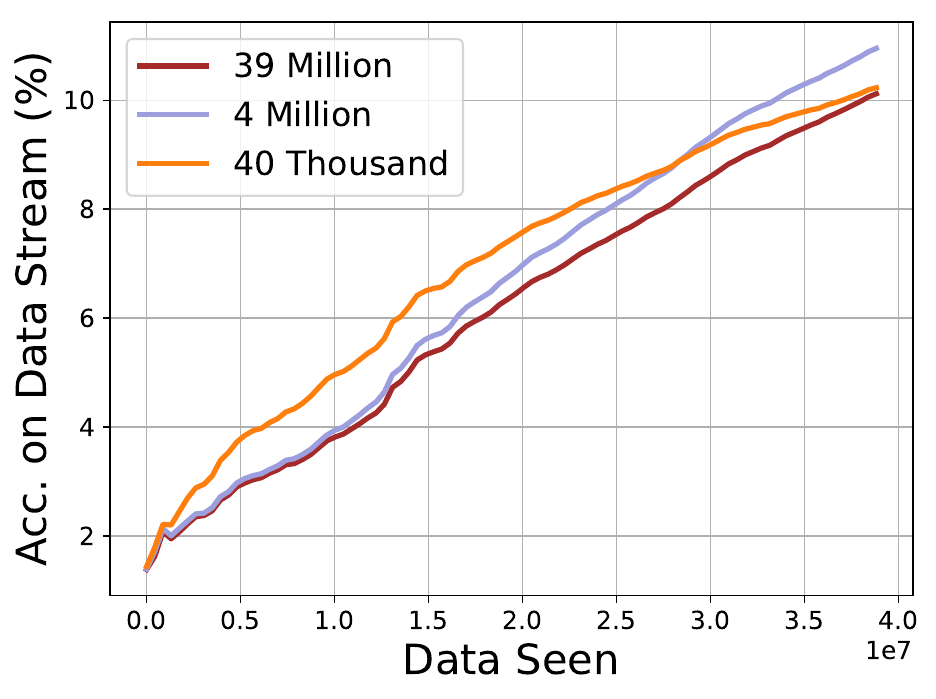}\label{fig:acc_new}}
	\subfigure[Training accuracy $\texttt{acc}_{\texttt{Rep}}$ on replay data]{\includegraphics[width=0.49\columnwidth]{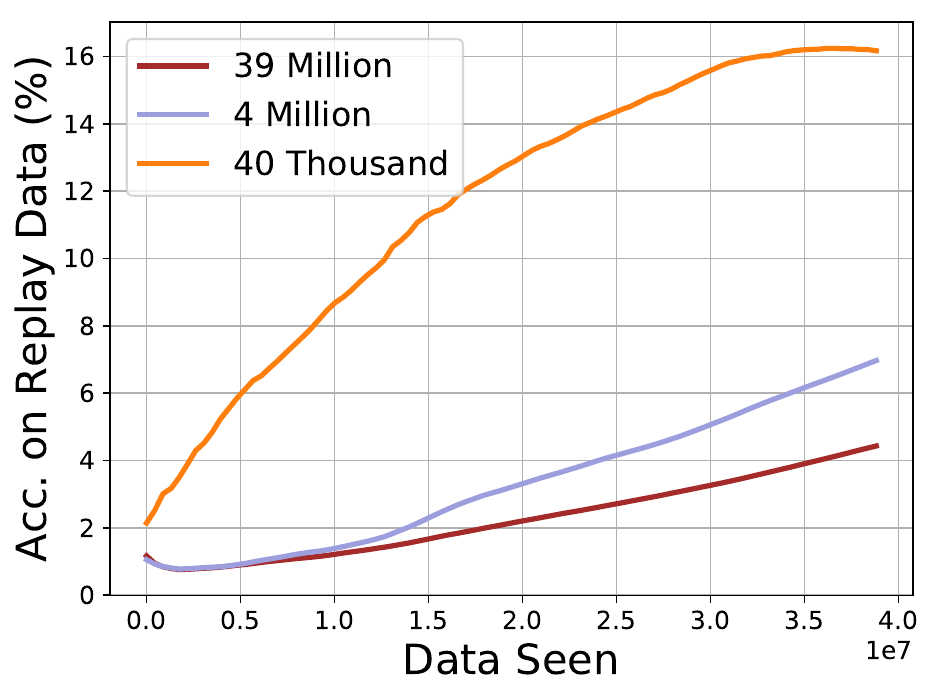}\label{fig:acc_old}}
	\caption{\textbf{The effect of replay buffer size $R$ on learning efficacy}. (a) The optimal replay buffer size changes over time. (b,c) Correlation between replay buffer size and training statistics. When $R$ was too small (40 thousand at late time steps), training accuracy $\texttt{acc}_{\texttt{Rep}} \ggreater \texttt{acc}_{\texttt{Stream}}$ on the data stream. When $R$ was too large (4 million and 39 million), $\texttt{acc}_{\texttt{Rep}} \llesser \texttt{acc}_{\texttt{Stream}}$.}
	\vspace{-1em}
\end{figure}

We report learning efficacy in Fig.~\ref{fig:varied_buf_size} for all models and conclude that the optimal buffer size
changes through time and the common choice of using the largest possible buffer is not always optimal. Interestingly, we
found that it is possible to
successfully adapt the buffer size by comparing the training accuracy on replay data and online data stream. We define training accuracy on the data stream at time $t$ as
\vspace{-3em}
\begin{align}\label{eq:NBAcc}
	\texttt{acc}_{\texttt{Stream}}(t) = \frac{1}{t} \sum_{s = 1}^{t} \texttt{acc}(Y_s, h(\XX_s; \btheta_s)),
\end{align}
where $\XX_s$, $Y_s$ are the training data and labels from the data stream at time $s$. Similarly, we define
the training accuracy on replay data at time $t$ as
\begin{align}\label{eq:RepAcc}
\texttt{acc}_{\texttt{Rep}}(t) = \frac{1}{t} \sum_{s = 1}^{t} \texttt{acc}(Y^{\text{Rep}}_s, h(\XX_s^{\text{Rep}}; \btheta_s)),
\end{align}
where $\XX_s^{\text{Rep}}$, $Y_s^{\text{Rep}}$ are the training data and labels sampled from the replay buffer.


We can see from Figs.~\ref{fig:acc_new} and~\ref{fig:acc_old} that when $R$ was too large (4 million and 39 million), the accuracy on replay
data $\texttt{acc}_{\texttt{Rep}}$ was much lower than the accuracy on the data stream $\texttt{acc}_{\texttt{Stream}}$,
making the model focus less on adaptation. When $R$ was too small (40 thousand at late training stages),
$\texttt{acc}_{\texttt{Rep}}$ was much higher than $\texttt{acc}_{\texttt{Stream}}$. This overfitting on the replay
buffer harmed the replay-based regularization.

\begin{figure}[t]
	\subfigure[Average online accuracy ($\uparrow$)]{\includegraphics[width=0.48\columnwidth]{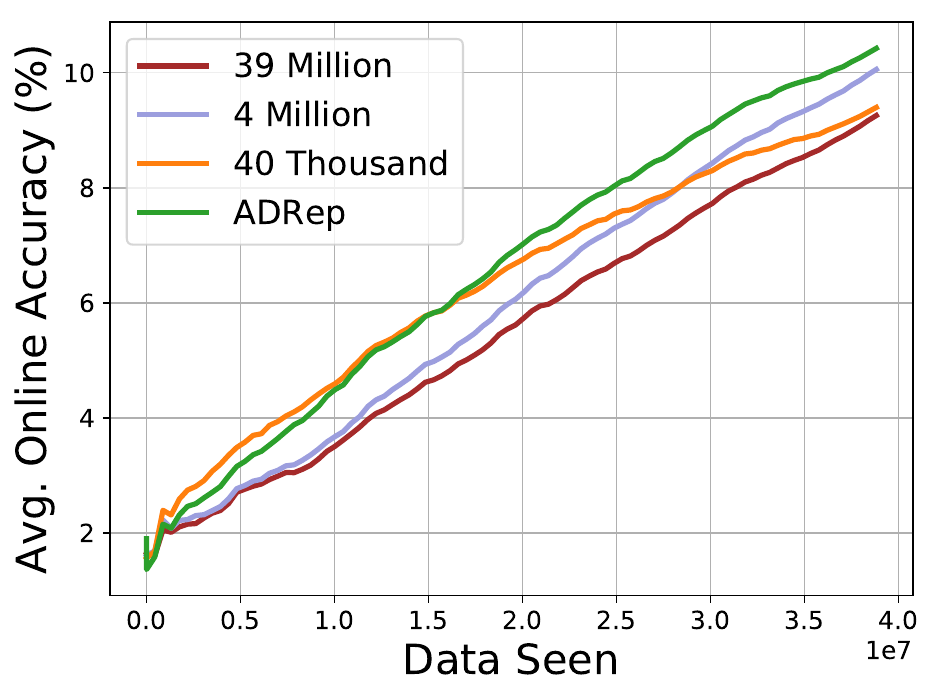}\label{fig:ADRep_online_fit}}
	\subfigure[Replay buffer sizes of ADRep over time]{\includegraphics[width=0.48\columnwidth]{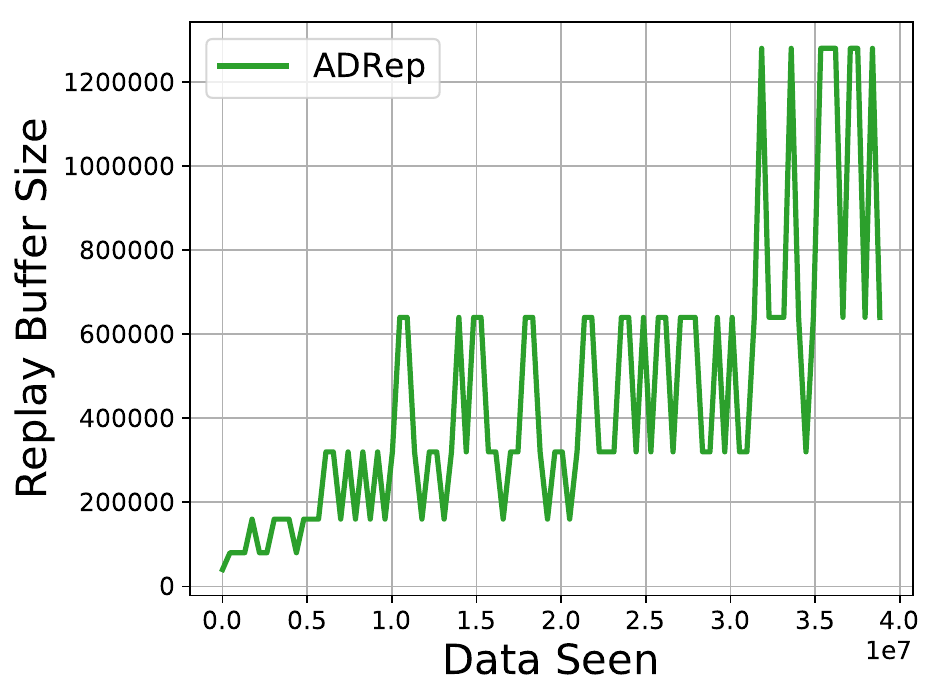}\label{fig:ADRep_buf_size}}
\caption{\textbf{ADRep vs.\ constant replay buffer sizes}. (a) ADRep provided near-optimal average online accuracy throughout training. (b) ADRep gradually increased the replay buffer size over time.}\label{fig:ADRep}
\vspace{-1em}
\end{figure}


\begin{figure}[t]
	\center
	\subfigure[Backward transfer at $H$ ($\uparrow$)]{\includegraphics[width=0.55\columnwidth]{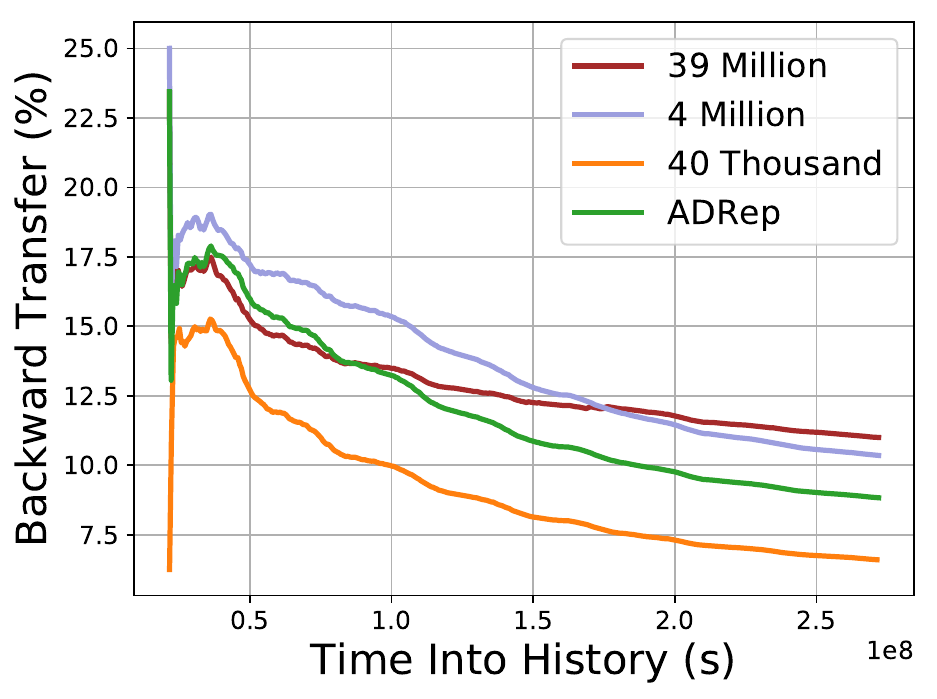}\label{fig:ADRep_BT_epoch90}}
	\subfigure[Backward transfer at $\nicefrac{2H}{3}$ ($\uparrow$)]{\includegraphics[width=0.49\columnwidth]{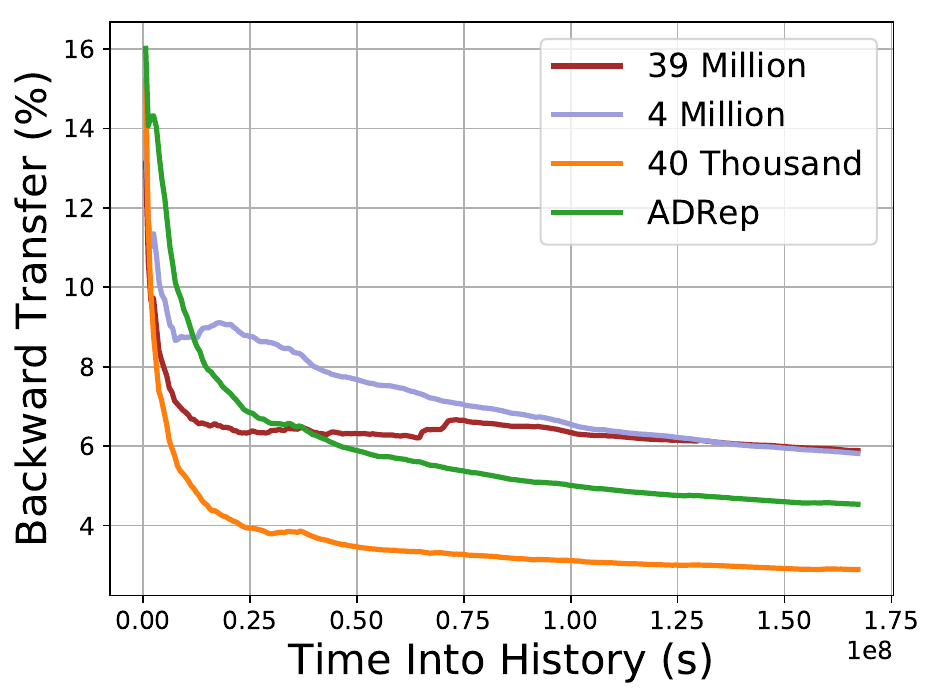}\label{fig:ADRep_BT_epoch60}}
	\subfigure[Forward transfer at $\nicefrac{2H}{3}$ ($\uparrow$)]{\includegraphics[width=0.49\columnwidth]{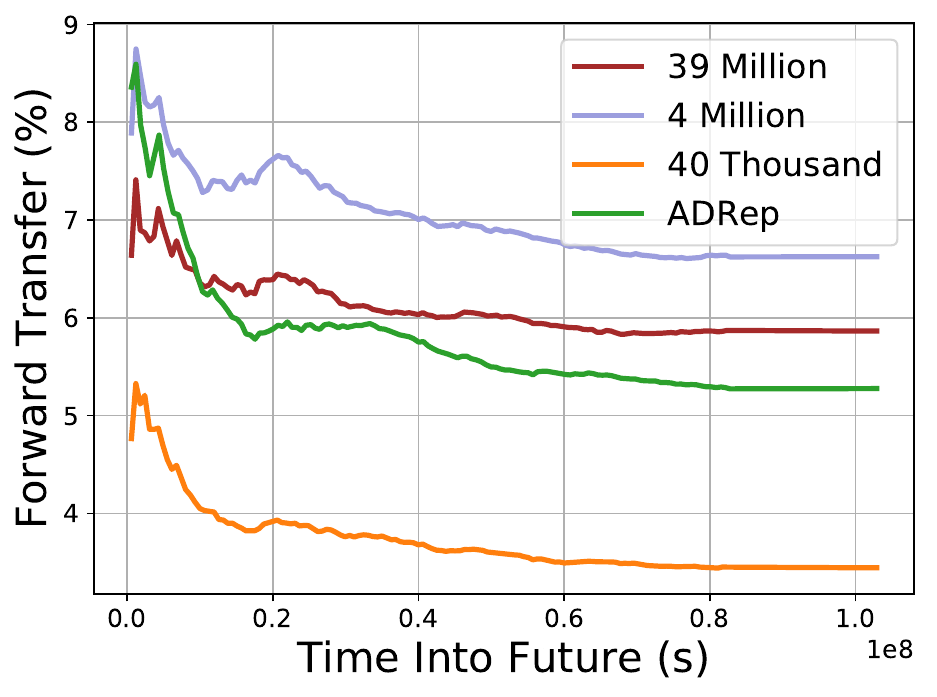}\label{fig:ADRep_FT_epoch60}}
	\caption{\textbf{Transfer results of different replay buffer sizes}. $H$ is the final time step. (a) Replay buffer size 39 million had comparable backward transfer to 4 million. (b,c) Replay buffer size 39 million had worse forward and backward transfer to 4 million. These results show that large replay buffer sizes are not always beneficial for transfer.}\label{fig:ADRep_Transfer}
	\vspace{-1.5em}
\end{figure}

	\begin{algorithm}[t]
	\begin{algorithmic}[1]
		\Require Update interval $N'$, initial replay buffer size $R$, difference threshold $\epsilon$, initial model $\btheta_1$.
		\State $\texttt{acc}_{\texttt{Stream}}, \texttt{acc}_{\texttt{Rep}} \leftarrow 0$; $k \leftarrow 1$
		\For{$t \in \{1,2,...\}$}
		\State update model with replay buffer size of $R$.
		\State $\texttt{acc}_{\texttt{Stream}} \leftarrow \frac{(k-1)\texttt{acc}_{\texttt{Stream}} + \texttt{acc}(Y_t, h(\XX_t; \btheta_t))}{k}$ \label{line:acc_stream}
		\State $\texttt{acc}_{\texttt{Rep}} \leftarrow \frac{(k-1)\texttt{acc}_{\texttt{Rep}} + \texttt{acc}(Y_t^{\text{Rep}}, h(\XX_t^{{\text{Rep}}}; \btheta_t))}{k}$\label{line:acc_rep}
		\State $k\leftarrow k+1$
		\If{$t \mod N' = 0$}
		\State $k \leftarrow 1$
		\State \textbf{If} $\texttt{acc}_{\texttt{Stream}}> \texttt{acc}_{\texttt{Rep}} + \epsilon$ \textbf{then} $R \leftarrow \frac{R}{2}$
		\State \textbf{If} $\texttt{acc}_{\texttt{Stream}} < \texttt{acc}_{\texttt{Rep}} - \epsilon$ \textbf{then} $R \leftarrow 2R$
		\EndIf
		\EndFor
	\end{algorithmic}
	\caption{Adaptive replay buffer size (ADRep)}\label{alg:ADRep}
\end{algorithm}

Based on this observation, we propose to adapt the buffer size $R$ over time by comparing $\texttt{acc}_{\texttt{Stream}}$ and
$\texttt{acc}_{\texttt{Rep}}$. We call this adaptive algorithm ADRep and specify it in Alg.~\ref{alg:ADRep}. Every
 $N'^{th}$ iteration, if $\texttt{acc}_{\texttt{Stream}} > \texttt{acc}_{\texttt{Rep}} + \epsilon$, we decrease $R$.
 And if $\texttt{acc}_{\texttt{Stream}} < \texttt{acc}_{\texttt{Rep}} - \epsilon$, we increase $R$. 
 We observed that
 ADRep is not sensitive to the hyperparameter choices, and we choose $\epsilon = 0.5\%$ and $N' \approx$ 40K.
 Fig.~\ref{fig:ADRep} shows the result of ADRep. From Fig.~\ref{fig:ADRep_online_fit}, we see that ADRep shows significantly
 superior performance throughout training. More importantly, Fig.~\ref{fig:ADRep_buf_size} shows that ADRep was able to
 gradually adapt $R$ over time.  Note that ADRep only requires to compute and compare online training statistics to adapt
 replay buffer sizes, which requires almost no extra computational/storage budget.

	In terms of transfer, Fig.~\ref{fig:ADRep_Transfer} plots the transfer results of all models. We see that $R = 4$ million
	was better than ${R=39 \text{ million}}$ at time $\nicefrac{2H}{3}$ and comparable at time $H$. Therefore, as with learning efficacy,
	larger $R$ is not always beneficial for transfer. ADRep did not perform well with respect to transfer in comparison to other strategies.
	
\vspace{-1em}
	
	\paragraph{Summary of findings.} (1) Larger replay buffer sizes are not always better both in terms of learning efficacy and
	transfer. (2) In terms of learning efficacy, the proposed adaptation strategy is successful, with almost no extra computation cost.

	\subsection{Batch Size Analysis}
	\label{sec:BS}


		\begin{figure}
		\subfigure[Training loss ($\downarrow$)]{\includegraphics[width=0.49\columnwidth]{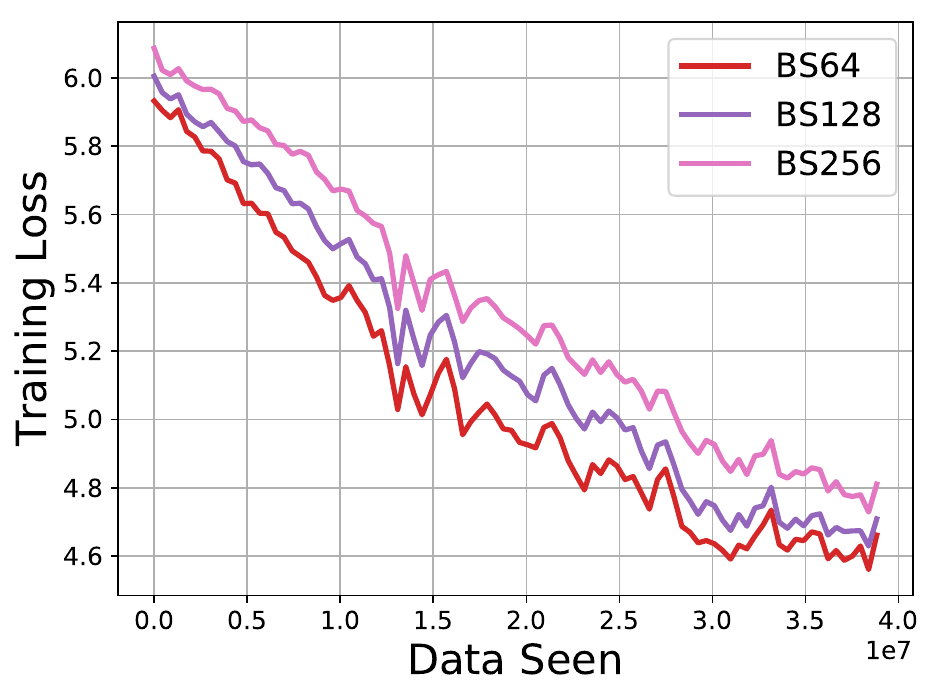}\label{fig:BS_train_loss}}
		\subfigure[Average online accuracy ($\uparrow$)]{\includegraphics[width=0.49\columnwidth]{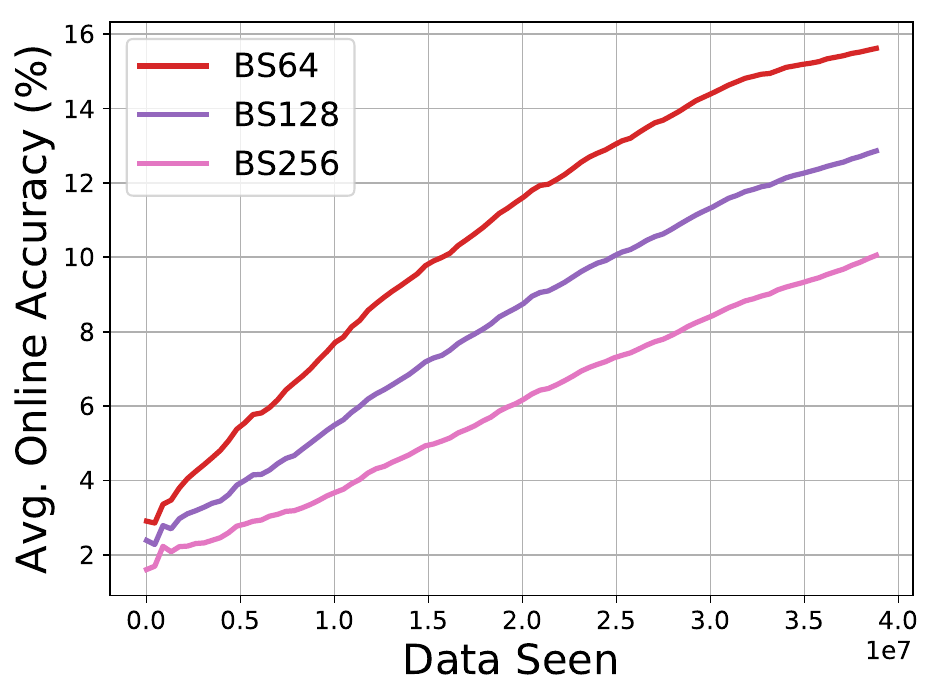}\label{fig:BS_online_fit}}
		\subfigure[Forward transfer at $\nicefrac{2H}{3}$ ($\uparrow$)]{\includegraphics[width=0.49\columnwidth]{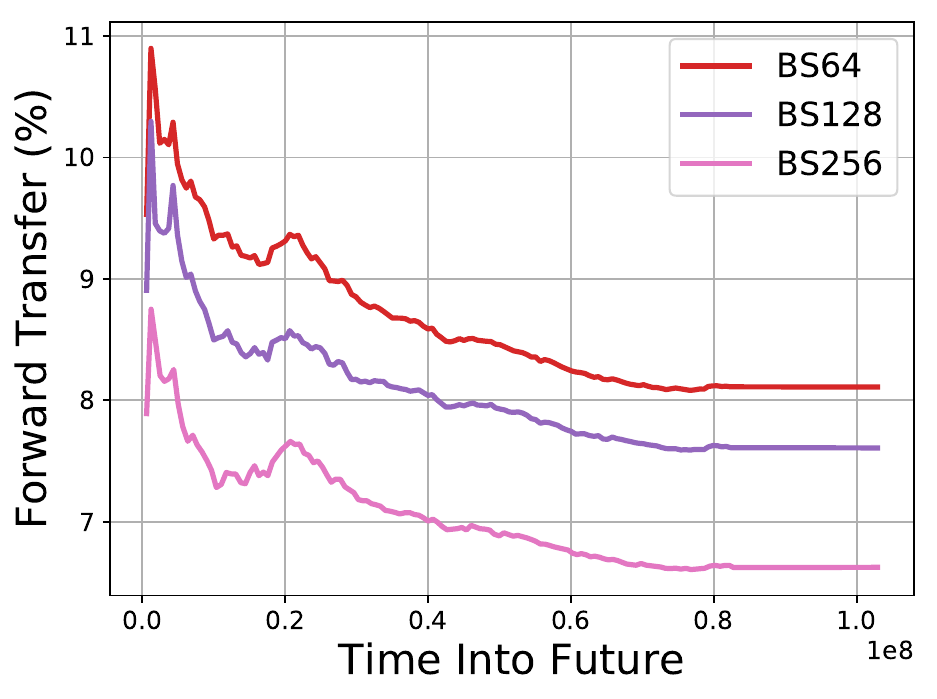}\label{fig:BS_FT_epoch30}}
		\subfigure[Backward transfer at $\nicefrac{2H}{3}$ ($\uparrow$)]{\includegraphics[width=0.49\columnwidth]{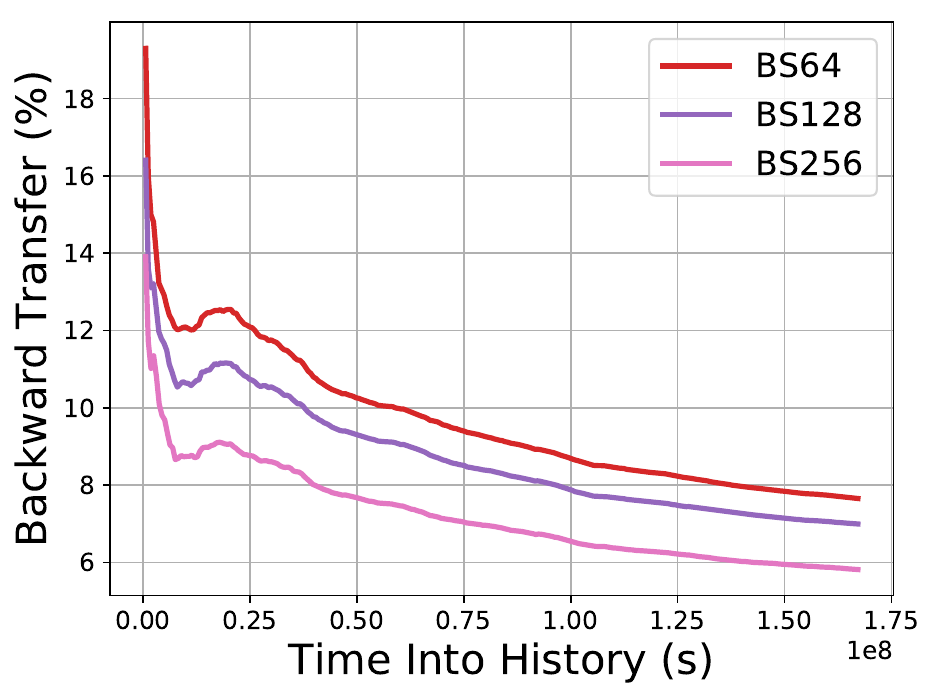}\label{fig:BS_BT_epoch30}}
		\caption{\textbf{The effect of batch size}. $H$ is the final time step. Increasing batch size has consistent and strong negative effect on the training loss (a) and all performance metrics, including average online accuracy (b), forward transfer (c), and backward transfer (d).}\label{fig:BS}
		\vspace{-1.5em}
	\end{figure}

	The concept of batch size is difficult to define in online
	continual learning. The natural setting is a batch size of $1$ album (1.157 on average on CLOC), but $\frac{39\text{ million}}{1.157} \approx 34\text{ million}$ iterations of SGD in order to evaluate a model are not practically feasible. Moreover, supervised learning practice suggests that mini-batching is beneficial, especially for saving training time. To separate evaluation from implementation, we evaluate online accuracy using the batch size of single album, but the training process is free to use any batch size.

    Analysis in supervised learning~\cite{goyal2017accurate}
    suggests that learning rates and batch sizes should be scaled similarly. Specifically, training with batch size $F\cdot B_0$ and learning rate $F\cdot l_0$ will have similar learning dynamics to training with batch size $B_0$ and learning rate $l_0$, as long as $F\cdot B_0$ is not too large, e.g., $\leq$ 4 thousand. This behavior is repeatable for CLOC when we use supervised learning on shuffled data. (See the supplement for details.)

    We similarly analyze the effect of batch sizes to OCL. We group $B$ consecutive samples into a batch
	and do training on each batch of data once we see $B$ images (Testing is always done with the newest model independently.). To analyze the effect, we train three models with $B$ set to 64, 128,
	and 256. For $B<256$, we reduce the learning rate to $\frac{B}{256}l$, where $l$ is the learning rate used for $B = 256$.
	We use a cosine schedule and a 4 million replay buffer here. Unlike in supervised learning,
	increasing $B$ (even by a small factor $F$) heavily hurts all performance metrics of OCL.

	As shown in Fig.~\ref{fig:BS}, the training loss (Fig.~\ref{fig:BS_train_loss}) as well as all performance metrics (Figs.~\ref{fig:BS_online_fit} to~\ref{fig:BS_BT_epoch30}) changed dramatically
    with $B$. Increasing $B$ had a consistent and strong negative effect on all metrics.
    This was true whether we use a replay buffer or not. The supplement provides similar plots for models trained without ER. This means that the effect of batch size is not an
	algorithm-specific result.

	In supervised learning, data is iid.\ and gradient estimates are unbiased regardless
	of batch sizes. Batch sizes only affect the variance. In OCL, gradients are not
	unbiased as the distribution is non-iid. With increasing batch sizes, the variance decreases but bias increases.
	Due to the differences between offline and online continual learning, gradient-based optimization requires additional
	care for online continual learning, further validating a benchmark such as CLOC.

	\vspace{-1em}
\paragraph{Summary of findings.} Unlike in supervised learning, mini-batching in OCL is not straightforward. Increasing the batch size, even by a small factor, has a strong negative effect on all performance metrics. This suggests that the smallest possible batch size should be used.

\subsection{Overall Performance}\label{sec:best_model}
		\begin{figure}
	\subfigure[Avg. online accuracy ($\uparrow$)]{\includegraphics[width=0.44\columnwidth]{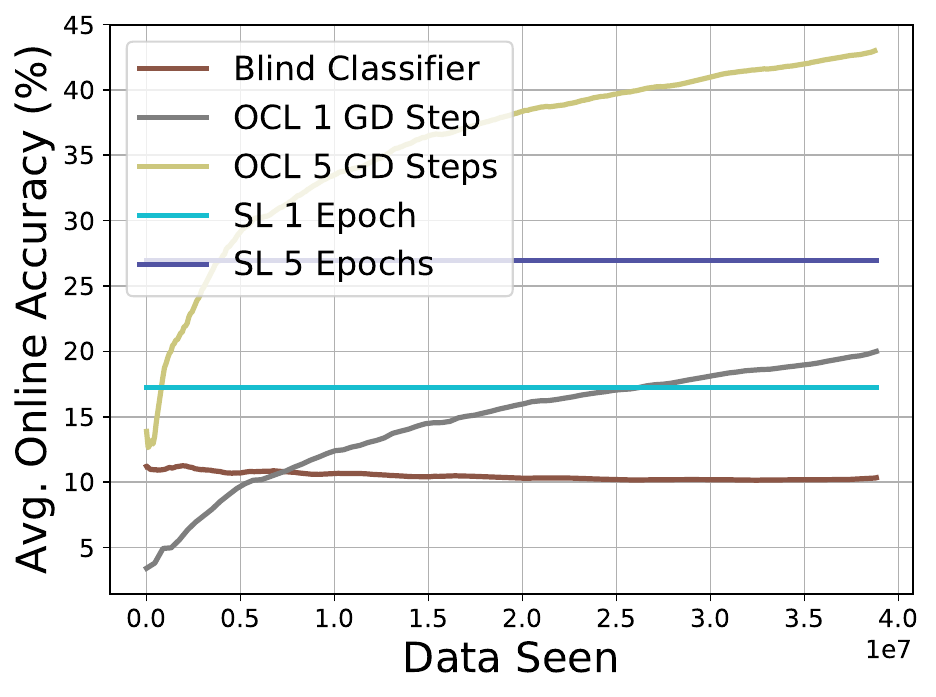}\label{fig:online_fit_best}}
	\subfigure[Top-1 predicted location eror ]{\includegraphics[width=0.55\columnwidth]{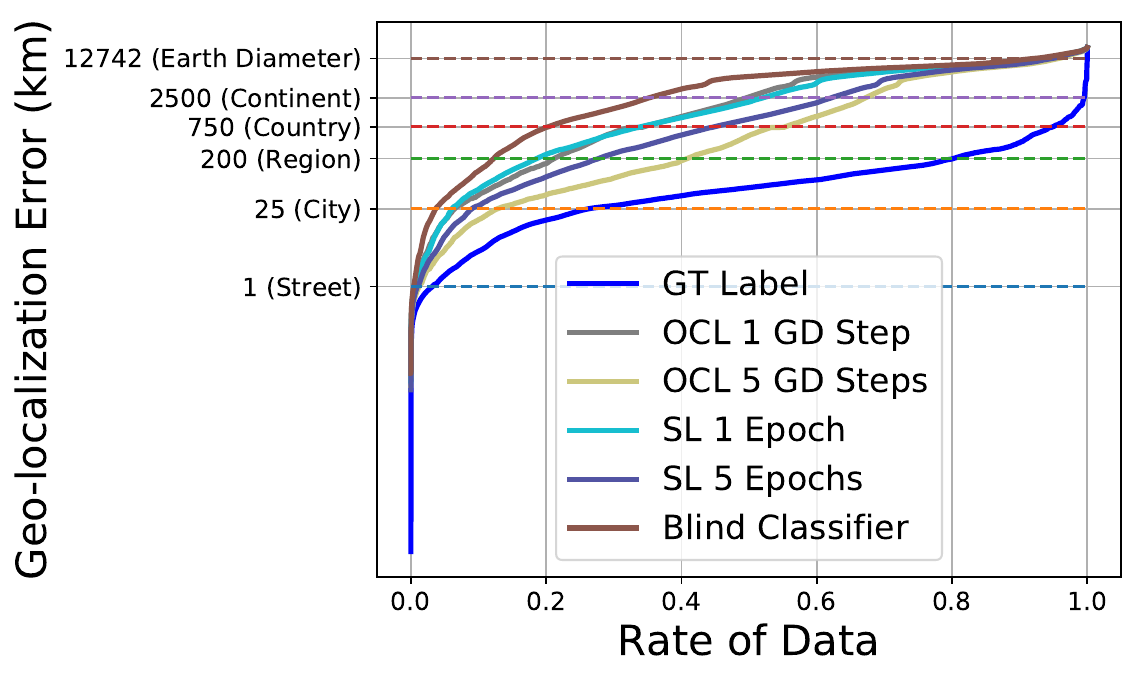}\label{fig:distance_best}}
	\caption{\textbf{OCL model vs.\ others}. SL refers to the result of supervised learning. "GD Step" means the number of gradient descent steps per mini-batch. (a) Average online accuracy of OCL vs. average online accuracy of the blind classifier and the validation accuracy of supervised models. The average online accuracy of OCL was comparable to the validation accuracy of supervised learning models. (b) Geolocalization error of top-1 OCL predictions. Bottom right is better.}\label{fig:BestModel}
	\vspace{-1.5em}
\end{figure}

In this section, we develop an online continual learning strategy that combines the presented findings and evaluate its
performance. Specifically, we utilize the adaptive learning rate schedule from PoLRS, adaptive buffer size from ADRep,
 and small batch sizes. We set the batch size to 64, which is the smallest value such that training on the whole dataset
 fits within a reasonable amount of time.

To analyze this model, we compare its average online accuracy against the performance of two other classifiers. The first one is the average online
accuracy of a "blind" classifier, which does not use input images. The blind classifier only uses historical labels to predict the labels
 of future examples, without using the actual images. Specifically, we use the most frequent labels that appeared in the
 previous $10$ images (tuned to achieve the maximum average online accuracy). The second
 classifier is the supervised learning model. We train supervised learning models with two time budgets, one for one
 epoch on the whole training set, and the other for five epochs.
 We compare the average online accuracy of OCL to the average validation accuracy of the supervised learning models. To examine the effect of the training budget for OCL, we also trained two OCL models with respectively one and five gradient descent steps per mini-batch.

 The results are summarized in Fig.~\ref{fig:BestModel}.
 Fig.~\ref{fig:online_fit_best} indicates that the blind classifier had significantly better performance than chance, which would have $\frac{1}{712}$ accuracy. In turn, the average online accuracy of OCL rose significantly beyond the accuracy of the blind classifier. This indicates that the model successfully learned over time. Somewhat surprisingly and promisingly, the online accuracy of OCL
 was better than the validation accuracy of supervised learning models given similar budgets. This indicates that learning efficacy and information retention are conflicting in some aspects, i.e., optimizing one objective may hurt the other. Hence, it is important to choose the right optimization objective in practice. 


  We also plot the distance between the location of each image and the location represented
  by the top-1 prediction of OCL (Fig.~\ref{fig:distance_best}).
  Similar to the supervised learning results of PlaNet~\cite{weyand2016planet}, most of the predictions were far from the actual location of the image
  because of the inherent difficulty of visual geolocalization. Nonetheless, OCL was better than supervised learning in
  terms of the geolocalization error.

\section{Conclusion}\label{sec:Conclusion}
We studied the problem of online continual learning with visual data. Using images with time
stamps and geolocation tags, we proposed a new benchmark with a large scale and
natural distribution shifts. We analyzed the effect of major optimization
decisions, including learning rates, replay buffer sizes, and batch sizes. We found that the ideal learning rates are different for learning efficacy and information retention, and proposed different schedules for different performance metrics. We also found that the common practice of using the maximum possible replay buffer size for experience replay is not always optimal. We proposed an online replay buffer size adaptation algorithm to improve learning efficacy. For batch sizes, we found that mini-batching in OCL is non-trivial. Unlike in supervised learning, where mini-batch SGD is the standard for parallelized training, increasing batch sizes, even by a small amount, severaly hurts both learning efficacy and information retention. Hence, the smallest possible batch size should be used for OCL. Using the proposed strategies, we were able to train an
OCL model with comparable online accuracy to the validation
accuracy of supervised learning models trained with similar budgets. 

\vspace{-1em}

\paragraph{Future directions.}
Many interesting future research directions have emerged from this work. For example, in Sec.~\ref{sec:best_model}, we see that the blind classifier can use the temporal coherence of the labels to achieve much better online performance than random guessing. Leveraging such temporal coherence to improve our model, which only uses a single image as input, is an interesting possibility. Meanwhile, this work mainly studies ``supervised" OCL, where the label of every example is observable. Extending the study to ``semi-supervised" or ``self-supervised" online visual continual learning is also an interesting topic.

\clearpage

\begin{center}
{\Large \textbf{Suppliment}}
\end{center}

\appendix

\begin{abstract}
	We present additional experimental results in this supplement, which we skipped in the 
	main text due to space limitations. In Section~\ref{appd:FIFOvsReservoir},
	we present an additional study on the replay buffer update strategies. 
	In Section~\ref{appd:BatchSize}, we present an additional study on the impact of 
	batch sizes, specifically on the supervised learning counterpart of geolocalization, 
	and the setting of online continual learning without replay buffer.
\end{abstract}

\section{Effect of replay buffer update strategies}\label{appd:FIFOvsReservoir}

\begin{figure}[th]
	\subfigure[Average online accuracy ($\uparrow$). Replay buffer size = $40$ thousand.]{\includegraphics[width=0.49\columnwidth]{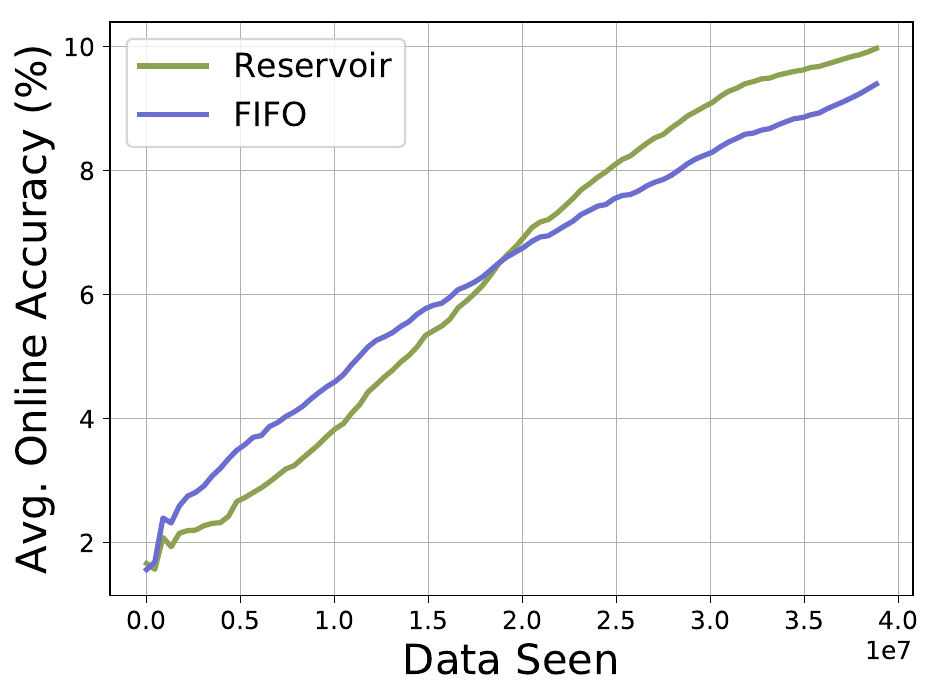}\label{fig:appd_online_fit_40K}}
	\subfigure[Backward transfer at final time step $H$ ($\uparrow$). Replay buffer size = $40$ thousand.]{\includegraphics[width=0.49\columnwidth]{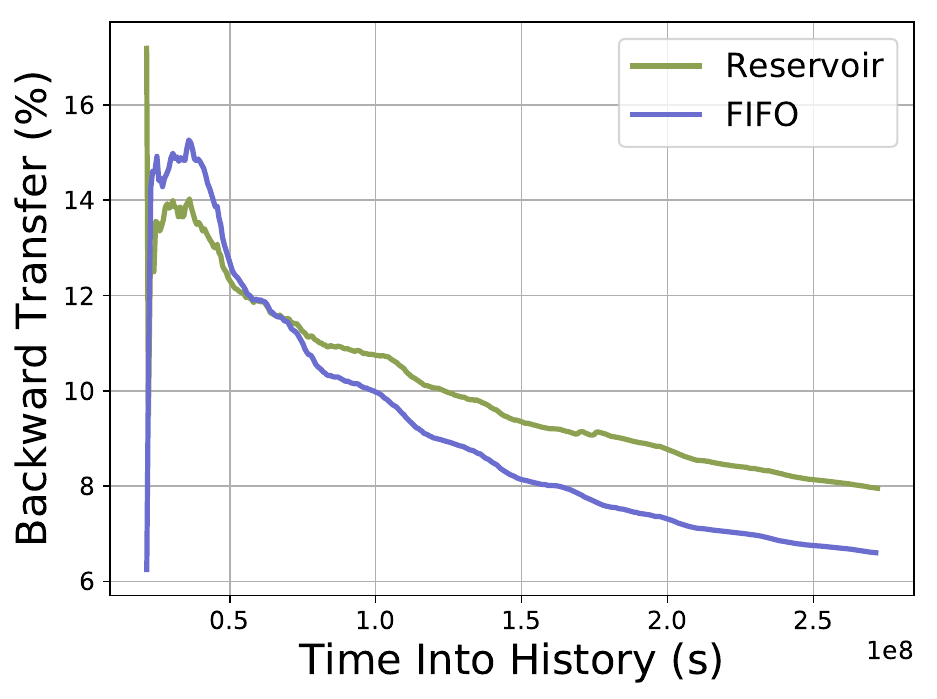}\label{fig:appd_BT_40K}}
	\subfigure[Average online accuracy ($\uparrow$). Replay buffer size = $4$ million.]{\includegraphics[width=0.49\columnwidth]{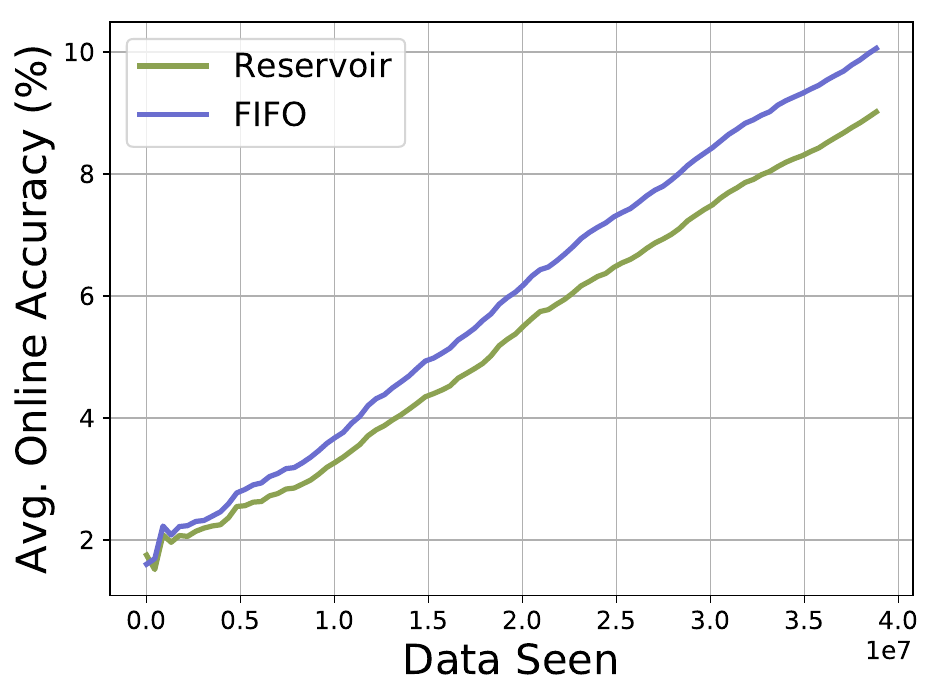}\label{fig:appd_online_fit_4M}}
	\subfigure[Backward transfer at final time step $H$ ($\uparrow$). Replay buffer size = $4$ million.]{\includegraphics[width=0.49\columnwidth]{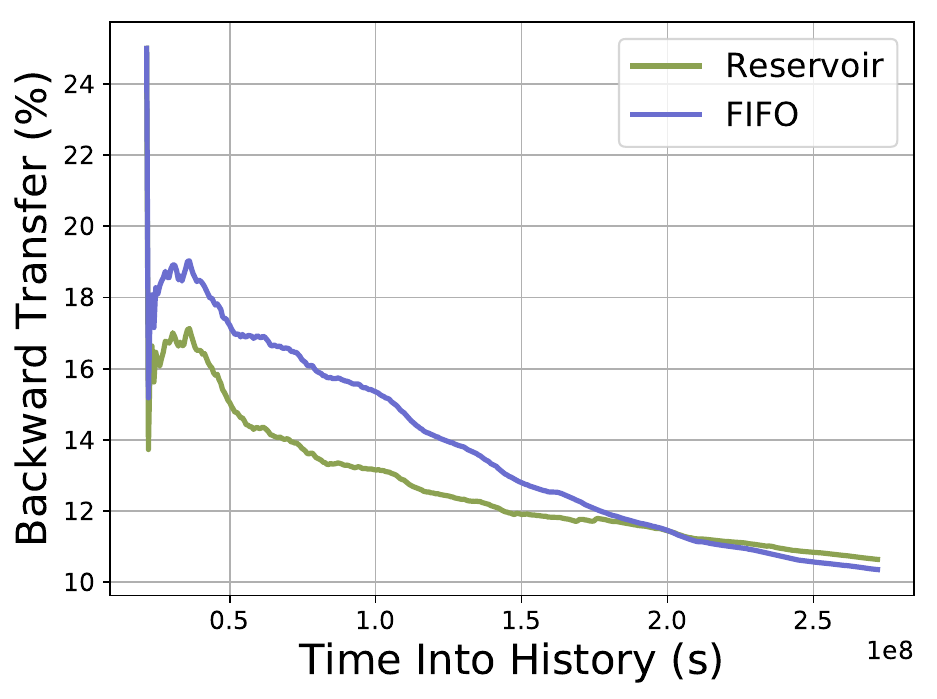}\label{fig:appd_BT_4M}}
	\caption{\textbf{FIFO vs reservoir buffer, with different replay buffer sizes.} The arrows in the caption of subfigures point towards better performance. Overall, FIFO and reservoir had comparable performance both in terms of learning efficacy and information retention. Reservoir performed slightly better (Fig.~\ref{fig:appd_online_fit_40K} and Fig.~\ref{fig:appd_BT_40K}) with $40$ thousand replay buffer size, but slightly worse (Fig.~\ref{fig:appd_online_fit_4M} and Fig.~\ref{fig:appd_BT_4M}) with $4$ million replay buffer size. }\label{fig:BufType}
\end{figure}

As mentioned in Sec.~5.1 of the main paper, we choose the First-In-First-Out (FIFO) 
buffer for experience replay. To demonstrate the effect of changing replay buffer update
 strategies, we compare FIFO with the reservoir replay buffer~\cite{chaudhry2019tiny}, 
 which has been applied to offline continual learning. Reservoir buffer constructs a 
 random iid. subset of the data seen so far using an iterative process. For the 
 $t^{th}$ example from the data stream, the reservoir buffer first generates a random 
 number $r$ uniformly sampled from $1$ to $t$. And if $r$ is not greater than the 
 replay buffer size $R$, the $r^{th}$ example in the replay buffer will be replaced by 
 the $t^{th}$ example from the data stream. 

In order to evaluate the impact of buffer strategy, we train models using FIFO and 
reservoir buffer. We use the cosine learning rate schedule, and two different replay 
buffer sizes, 40 thousand and 4 million. Fig.~\ref{fig:BufType} demonstrates the average
 online accuracy (Fig.~\ref{fig:appd_online_fit_40K} and~\ref{fig:appd_online_fit_4M}) 
 and backward trainsfer (Fig.~\ref{fig:appd_BT_40K} and~\ref{fig:appd_online_fit_4M}) 
 of models trained with different replay buffer strategies. We can see that FIFO and 
 reservoir had comparable performance, both in terms of learning efficacy, and 
 information retention. Since the impact of buffer strategy is not signficant, we 
 choose FIFO in our experiments due to its simplicity.

\section{Additional Analysis for Batch Size}
\label{appd:BatchSize}

\subsection{Batch size effect to supervised learning}
\begin{figure}[th]
		\centering
		\begin{tabular}{@{}c@{}}
			\includegraphics[width=0.8\columnwidth]{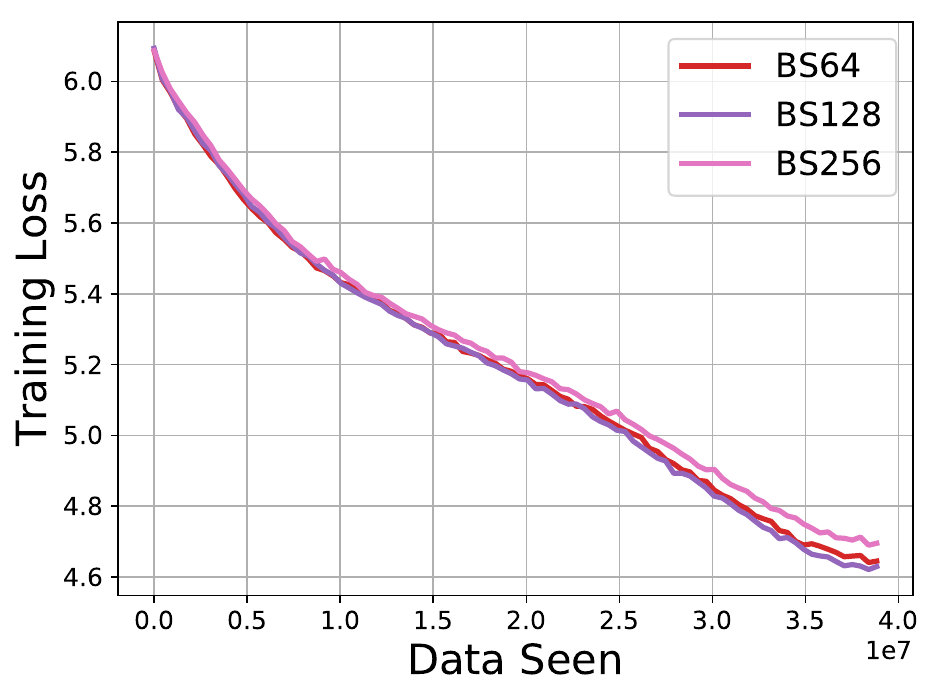} \\
			(a) Training loss \\ \\ \\
			\begin{tabular}{@{}cc@{}}
				\toprule
				Batch size & Validation accuracy \\
				\midrule
				64 & 16.61 \\
				128 & 17.07 \\
				256 & 17.21 \\
				\bottomrule						    
			\end{tabular} \\ 
		     \  \\
		  	(b) Validation accuracy ($\uparrow$)
		\end{tabular}
	\caption{\textbf{The effect of batch size to supervised leanring}. The arrow in the caption of (b) points towards better performace. Varying batch sizes had minor effect to the training loss (a) and validation accuracy (b).}\label{fig:BS_offline}
\end{figure}

In supervised learning, it is a common heuristics to multiply the batch size and 
learning rate by the same factor and recover similar learning dynamics. In contrast, 
our results show a strong negative effect with increasing batch sizes in online 
continual learning (OCL). We ask the question, is this unexpected behavior due to the 
data or the nature of OCL? We train supervised learning models with batch sizes of 64, 
128, and 256, respectively, using shuffled data to answer this question. Moreover, we 
set the learning rates to 0.0125, 0.025, and 0.05, respectively. All models are trained 
on our dataset for one epoch. As shown in Fig.~\ref{fig:BS_offline}, all models had 
similar training loss curves (Fig.~\ref{fig:BS_offline}(a)) and average validation 
accuracy (Fig.~\ref{fig:BS_offline}(b)). Unlike in the OCL case, increasing batch 
sizes from 64 to 256 was not harmful to supervised learning. It even slightly improved 
the validation accuracy. Hence, the batch size effect is due to the nature of the OCL 
problem.

\subsection{Batch Size Effect Without Replay Buffer}

\begin{figure}[th]
	\subfigure[Training loss ($\downarrow$)]{\includegraphics[width=0.49\columnwidth]{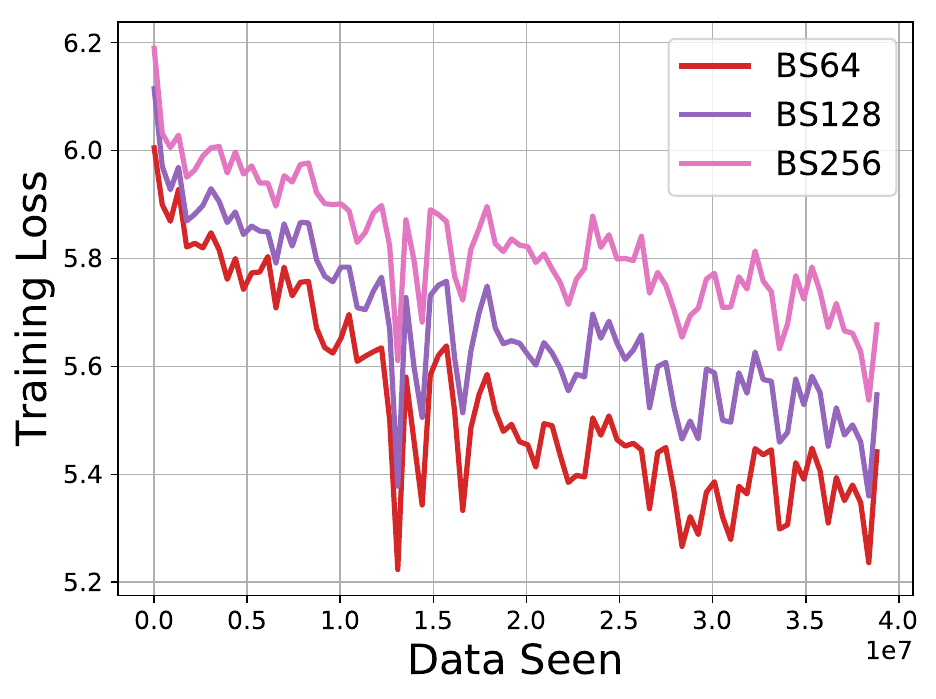}\label{fig:BS_train_loss_noRep}}
	\subfigure[Average online accuracy ($\uparrow$)]{\includegraphics[width=0.49\columnwidth]{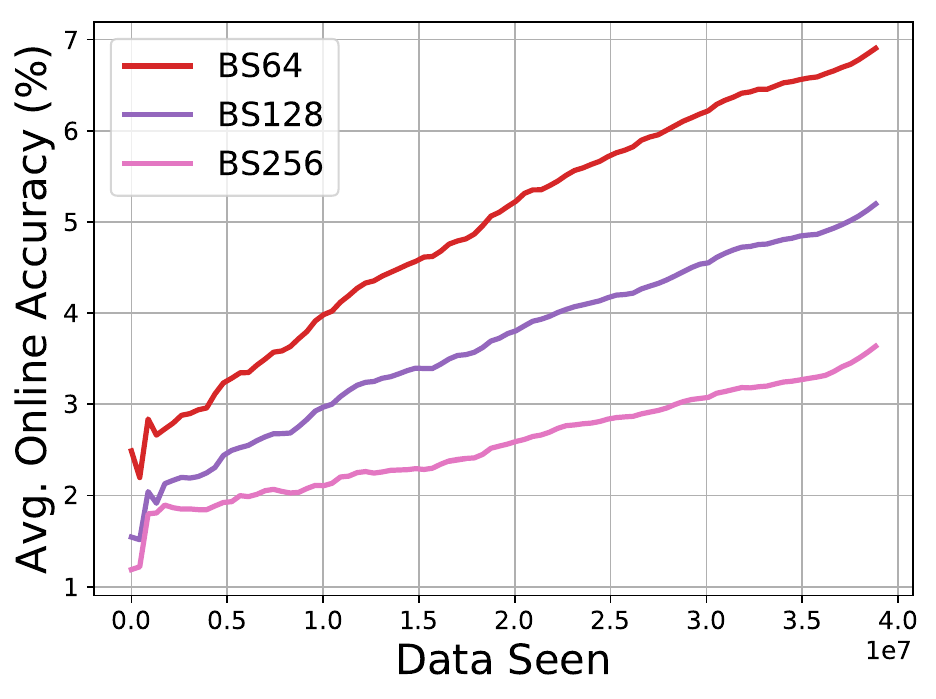}\label{fig:BS_online_fit_noRep}}	
	\subfigure[Backward transfer at $\frac{2H}{3}$ ($\uparrow$)]{\includegraphics[width=0.49\columnwidth]{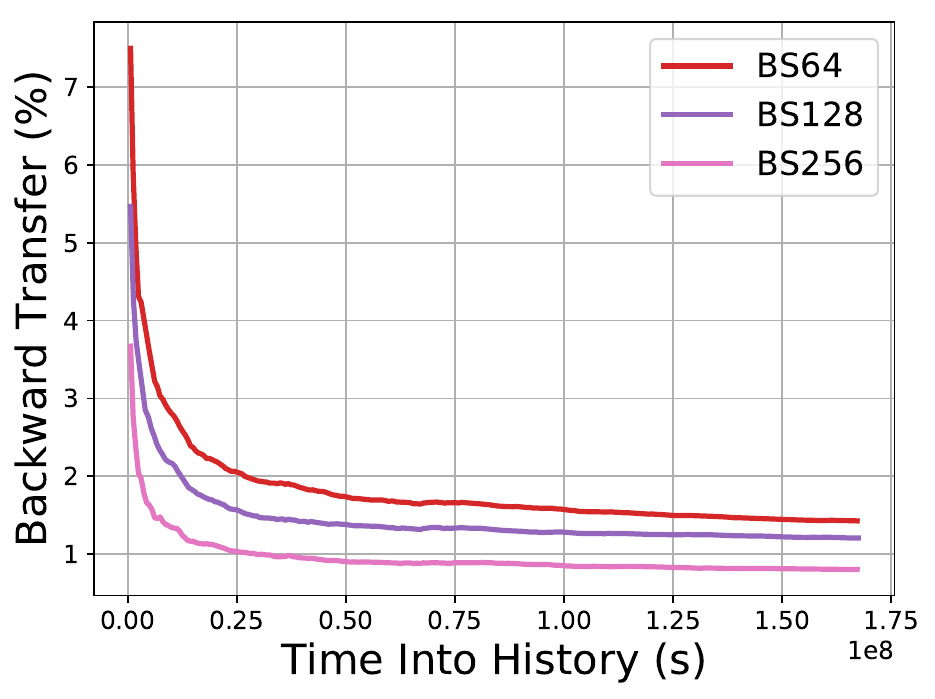}\label{fig:BS_BT_noRep}}
	\subfigure[Forward transfer at $\frac{2H}{3}$ ($\uparrow$)]{\includegraphics[width=0.49\columnwidth]{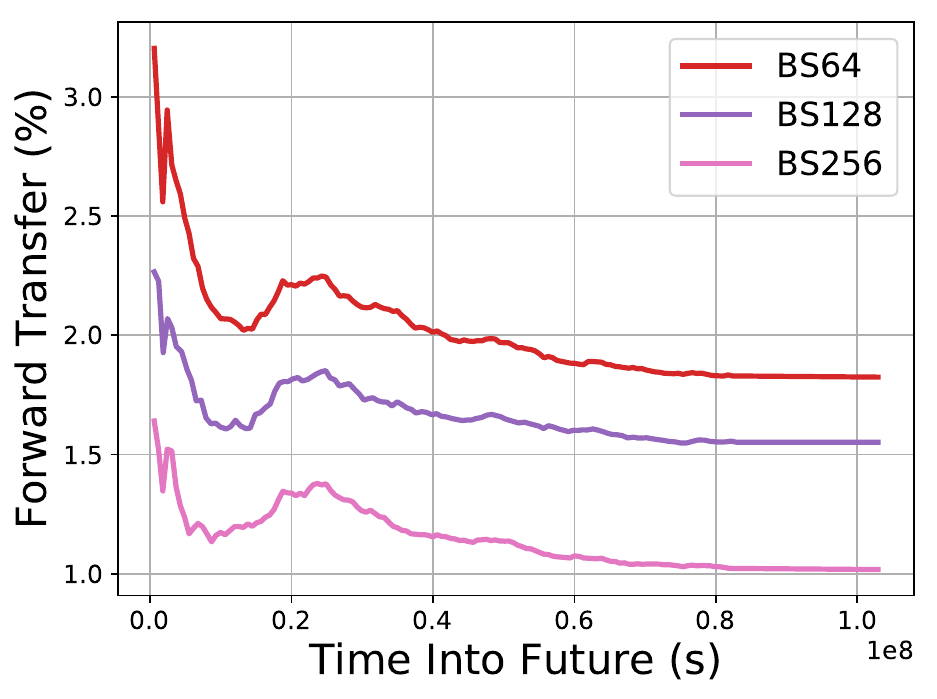}\label{fig:BS_FT_noRep}}	
	\caption{\textbf{The effect of batch size to OCL, without using replay buffer.} The arrows in the caption of subfigures point towards better performance. Similar to the result of the main paper, increasing batch sizes increased the training loss and hurts all performance metrics of OCL.}\label{fig:BS_no_rep}
\end{figure}

In the main paper, we analyze the effect of batch sizes to OCL, using models trained with 
experience replay. In order to validate whether the batch size effect we observed 
also exists without replay, we train OCL models using the same settings as in the 
batch size analysis of the main paper, except that all models are trained without 
experience replay. We plot the performance of trained models in Fig.~\ref{fig:BS_no_rep}. 
Similar to the main paper results, increasing batch sizes was harmful to the training 
loss and all performance metrics of OCL, even without experience replay. Hence, the 
batch size effect we observed for OCL is not algorithm-specific. 

\clearpage

{\small
	\bibliographystyle{ieee_fullname}
	\bibliography{egbib}
}

\end{document}